\documentclass[journal]{IEEEtran}
\ifCLASSINFOpdf
\else
\fi
\usepackage{graphicx}
\usepackage{array}
\usepackage{amsmath, bm}
\usepackage{booktabs, multirow, makecell} 
\usepackage{enumitem}

\usepackage{xcolor}
\usepackage[colorlinks]{hyperref} 
\usepackage{cite}

\usepackage{amsfonts}
\usepackage{colortbl}

\begin{document}
%
\title{MO R-CNN: Multispectral Oriented R-CNN for Object Detection in Remote Sensing Image}

%
%
%

\author{Leiyu Wang,
      	Biao Jin,
      	Feng Huang, 
      	Liqiong Chen,
      	Zhengyong Wang,
      	
      	Xiaohai He,~\IEEEmembership{Member,~IEEE},
      	Honggang Chen,~\IEEEmembership{Member,~IEEE}

\thanks{
	This work was supported in part by the Open Fund of Key Laboratory of the Ministry of Education on Artificial Intelligence in Equipment (Grant No. 2024-AAIE-KF04-03), 
	in part by the Science and Technology Projects of Xizang Autonomous Region, China (Grant No. XZ202501ZY0064), 
	in part by the Open Foundation of Yunnan Key Laboratory of Software Engineering (Grant No. 2023SE206), 
	and in part by the Fundamental Research Funds for the Central Universities (Grant No. SCU2023D062) \textit{(Corresponding authors: Honggang Chen; Zhengyong Wang.)}}

\thanks{Leiyu Wang is with the College of Electronics and Information Engineering, Sichuan University, Chengdu 610065, China (e-mail: 2022222050076@stu.scu.edu.cn).}
\thanks{Biao Jin is with the College of Electronics and Information Engineering, Sichuan University, Chengdu 610065, China (e-mail: 2024222055051@stu.scu.edu.cn).}
\thanks{Feng Huang is with the School of Mechanical Engineering and Automation, Fuzhou University, Fuzhou 350002, China (e-mail: huangf@fzu.edu.cn).}
\thanks{Liqiong Chen is with the School of Mechanical Engineering and Automation, Fuzhou University, Fuzhou 350002, China (e-mail: liqiongchen@fzu.edu.cn).}
\thanks{Zhengyong Wang is with the College of Electronics and Information Engineering, Sichuan University, Chengdu 610065, China (e-mail: wangzheny@scu.edu.cn).}
\thanks{Xiaohai He is with the College of Electronics and Information Engineering, Sichuan University, Chengdu 610065, China (e-mail: hxh@scu.edu.cn).}
\thanks{Honggang Chen is with the College of Electronics and Information Engineering, Sichuan University, Chengdu 610065, China, and also with the Yunnan Key Laboratory of Software Engineering, Yunnan University, Kunming 650600, China (e-mail: honggang\_chen@scu.edu.cn).}

}

%
%

\markboth{Journal of \LaTeX\ Class Files,~Vol.~14, No.~4, August~2025}%
{Shell \MakeLowercase{\textit{et al.}}: Bare Demo of IEEEtran.cls for IEEE Journals}
%



\maketitle 

\begin{abstract}
	Oriented object detection for multi-spectral imagery faces significant challenges due to differences both within and between modalities.
	Although existing methods have improved detection accuracy through complex network architectures, their high computational complexity and memory consumption severely restrict their performance.
	Motivated by the success of large kernel convolutions in remote sensing, we propose MO R-CNN, a lightweight framework for multi-spectral oriented detection featuring heterogeneous feature extraction network (HFEN), single modality supervision (SMS), and condition-based multimodal label fusion (CMLF).
	HFEN leverages inter-modal differences to adaptively align, merge, and enhance multi-modal features.
	SMS constrains multi-scale features and enables the model to learn from multiple modalities.
	CMLF fuses multi-modal labels based on specific rules, providing the model with a more robust and consistent supervisory signal.
	Experiments on the DroneVehicle, VEDAI and OGSOD datasets prove the superiority of our method.
	The source code is available at: https://github.com/Iwill-github/MORCNN.
\end{abstract}


%
\IEEEpeerreviewmaketitle

\section{Introduction}
%
%
%
%

%
%

\IEEEPARstart{M}{ultimodal} visible and infrared remote sensing image (VIRSI) oriented object detection is a core technology in modern remote sensing technology and intelligent detection systems.
It combines the rich color information and detailed clarity of visible remote sensing images (VRSI) with the thermal radiation information and all-weather observation capability of infrared remote sensing images (IRSI), and has been widely applied in various fields\cite{ wang2024multi, yue2025diffusion, wang2025cross} as a object detection method based on information fusion\cite{li2023cross, yuan2024improving,wang2025high}.
While significant research efforts by scholars have been dedicated to VIRSI multimodal oriented object detection, current VIRSI-oriented object detection algorithms still exhibit numerous limitations. 

\begin{figure}[!t]
	\centering
	\includegraphics[width=\columnwidth]{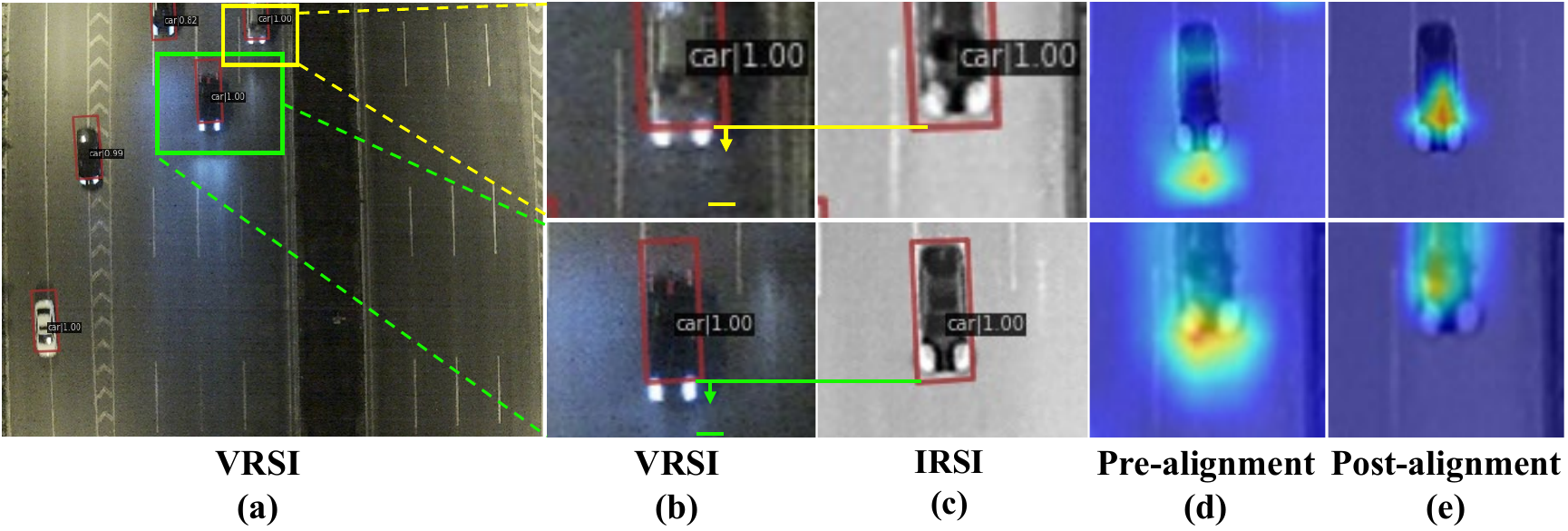}
	\caption{
		Instance position offset arising from multimodal acquisition time difference. 
		(a) shows the VRSI. (b) shows a local magnification of VRSI. (c) shows a local magnification of IRSI.
		(d) and (e) respectively represent the VRSI features before and after alignment.
		Due to the temporal asynchronicity of multispectral sensors, moving objects can result in displacement during cross-modal imaging.
	}
	\label{Instance_Position_Offset_Arising_from_Multimodal_Acquisition_Time_Difference}
\end{figure}

In terms of model design, the backbone in current multispectral algorithms for RSI are highly symmetrical, which simplifies the model but overlooks the heterogeneity among multispectral data.
This design limitation not only weakens the ability of model to specifically represent features of each spectral band but also can lead to the attenuation of spectrally specific information during feature fusion.
Furthermore, due to differences in sensor hardware and constraints in acquisition mechanisms, there exists a non-negligible temporal asynchrony between visible and infrared modalities. 
Spatial displacement of moving objects caused by asynchronous acquisition intervals can lead to misalignment of cross-modal features in the spatial dimension, as
shown in Fig. \ref{Instance_Position_Offset_Arising_from_Multimodal_Acquisition_Time_Difference}.
The traditional static feature fusion paradigm, such as channel concatenation or weighted averaging, lacks the capability for cross-modal semantic alignment, making it prone to feature mismatch when building cross-modal associations.
The existing dynamic feature fusion methods, such as transformer-based cross-attention, can achieve high-precision feature alignment, but they significantly increase the parameter and computational overhead. Furthermore, the subsequent addition of multi-branch detection heads exacerbates the issue of parameter redundancy, severely limiting the practical deployment feasibility of these models in memory-constrained scenarios.

\begin{figure}[!t]
	\centering
	\includegraphics[width=\columnwidth]{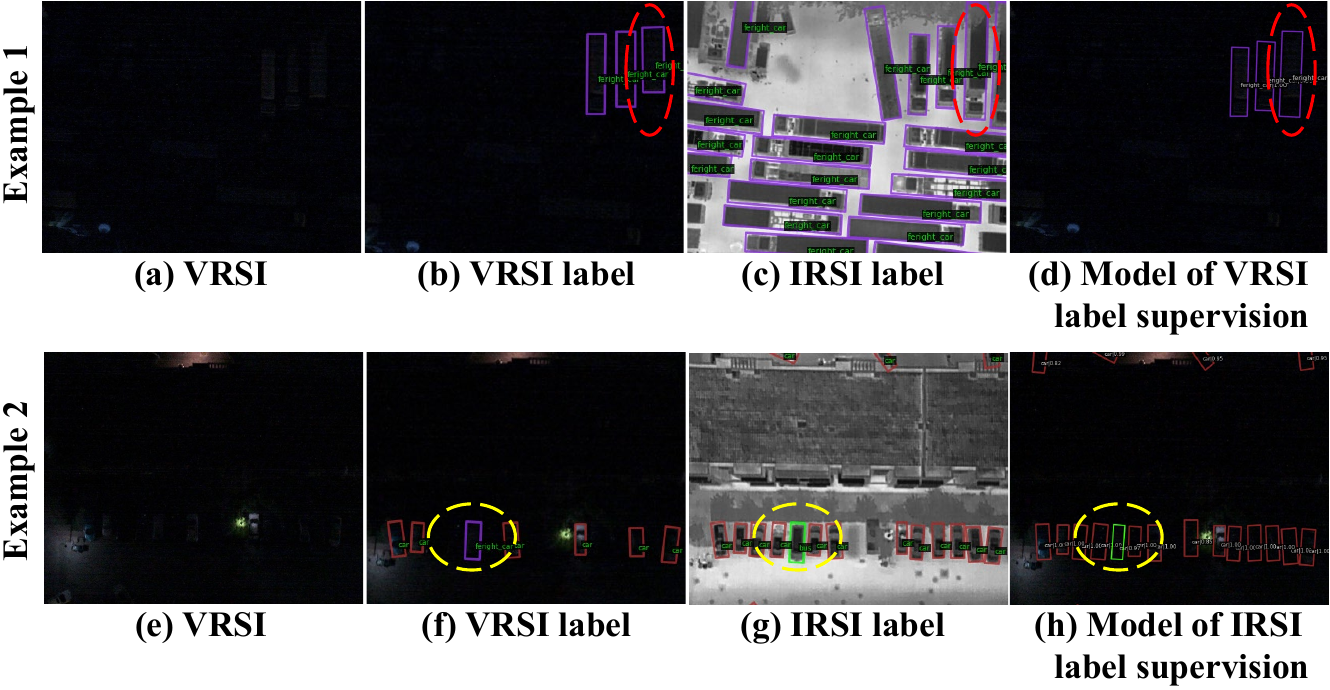}
	\caption{
		Shortcomings of training multimodal models with single modal labels. 
		Elliptical dashed boxes of the same color indicate the same target. 
		Rectangular boxes of the same color represent targets of the same class.
		(a) and (e) show the VRSI.
		(b) and (f) indicate the visualization of VRSI labels overlaid on the VRSI.
		(c) and (g) show the IRSI labels visualized on the IRSI.
		(d) and (h) represent the inference results of a multi-spectral input model supervised only by single-spectral labels.
	}
	\label{Shortcomings_of_Training_Multimodal_Models_with_Single_Modal_Labels}
\end{figure}

In terms of model training, existing studies typically employ single-modal labels for supervised learning, but this approach overlooks the issue of cross-modal annotation inconsistency.
Some datasets (e.g., VEDAI\cite{razakarivony2016vehicle} and OGSOD\cite{wang2023category}) maintain consistent annotations across different modalities through rigorous registration, thereby supporting supervised training with single-modality labels.
However, in more practically applicable scenarios, such as the DroneVehicle\cite{sun2022drone}, there are often discrepancies in bounding boxes and categories between different modal annotations due to sensor differences and inconsistent labeling.
As shown in Fig. \ref{Shortcomings_of_Training_Multimodal_Models_with_Single_Modal_Labels}, cross-modal annotation inconsistency problem leads to the model optimization being biased toward single-modality annotations, ultimately constraining the performance ceiling of multimodal fusion-based detection.

This paper proposes some innovative solutions to address the above issues.
Specifically, to address the heterogeneity in multi-spectral data, we propose heterogeneous feature extraction network (HFEN).
The network uses a heterogeneous dual stream LSK (HDS-LSK) for differentiated feature extraction of visible/infrared spectra, employs a selective multimodal feature fusion (SMFF) module for cross-modal feature alignment and fusion, and enhances the fused features using a residual feature augmentation (RFA) module.
HFEN effectively suppresses the attenuation of spectrally specific information while reducing computational costs.
Then, to optimize the redundant inference design of the three-branch classification and regression heads in existing multi-modal algorithms, single modality supervision (SMS) is proposed. SMS constrains multi-scale features while also endowing the model with the ability to learn from multiple modalities.
Finally, to address cross-modal annotation inconsistencies in multi-spectral datasets, condition-based multimodal label fusion (CMLF) is proposed. It fuses bi-modal labels based on specific rules, providing the model with a more robust and consistent supervisory signal.

In summary, our contributions are listed as follows:
\setlist[enumerate]{itemsep=1pt, topsep=1pt, parsep=1pt} 
\begin{enumerate}
	\item 
	We propose a lightweight framework for multi-spectral oriented detection, termed MO R-CNN. MO R-CNN employs HFEN for differentiated feature extraction, SMFF for cross-modal feature alignment and fusion, and RFA for multi-scale feature enhancement.
	\item 
	We propose a novel multi-spectral training strategy called SMS, which constrains multi-scale features while also endowing the model with the ability to learn from multiple modalities.
	\item 
	We propose the CMLF, which fuses multimodal labels through specific rules to alleviate the problem of inconsistent bimodal labels in practical scenarios from a training strategy perspective.
	\item 
	We conducted experiments on DroneVehicle \cite{sun2022drone}, VEDAI \cite{razakarivony2016vehicle} and OGSOD \cite{wang2023category} datasets. MO R-CNN achieves the lowest parameter while still surpassing existing methods in detection accuracy, thus establishing a new baseline for multi-spectral oriented detection.
\end{enumerate}

\begin{figure*}[!t]
	\centering
	\includegraphics[width=0.9\textwidth]{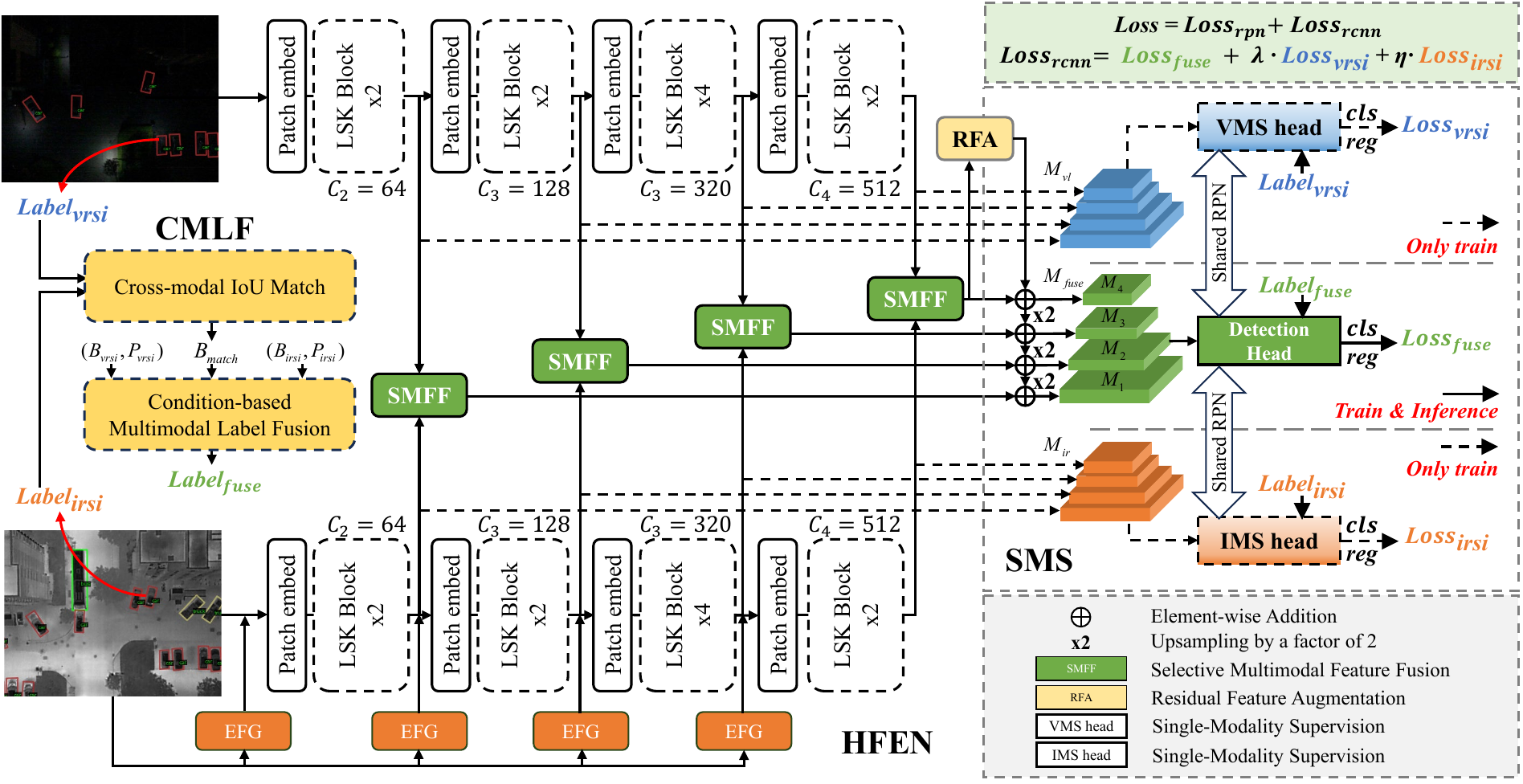}
	\caption{
		Overall framework of MO R-CNN. 
		The framework consists of three main components: the heterogeneous feature extraction network (HFEN), the single modality supervision (SMS), and the condition-based multimodal label fusion (CMLF).
		In Section A, we will focus on introducing the heterogeneous characteristics of HFEN. Then, in Section B, we will describe the proposed SMS strategy and discuss the loss function. Finally, the label fusion rules specified by CMLF will be presented in Section C.
	}
	\label{MO_R-CNN}
\end{figure*}

\section{Related Works}


\subsection{Multimodal RSI Oriented Object Detection}

Owing to the strong nonlinear fitting abilities of deep neural networks, deep learning has achieved significant progress in visual tasks. Earlier studies \cite{wagner2016multispectral,zhang2019cross,zhang2021weakly,zhou2020improving} concentrated on optimizing multi-spectral pedestrian detection algorithms from conventional viewpoints.
In recent years, the research community has expanded this area into oriented detection scenarios for RSI and has successively constructed multiple multi-modal remote sensing datasets.
Among these, VEDAI \cite{razakarivony2016vehicle}, OGSOD \cite{wang2023category} and DroneVehicle \cite{sun2022drone} have been established as benchmark datasets in the field of multi-modal RSI detection by providing large-scale registered RGB-IR image pairs and detailed annotations. This has inspired numerous scholars to engage in research in this area \cite{sun2022drone,yuan2022translation,yuan2024c,yuan2024improving,qingyun2022cross,zhang2023superyolo, wang2024yolofiv, zhang2024e2e, zhou2024dmm, bao2025dual}.
Notably, UA-CMDet \cite{sun2022drone} introduces an uncertainty-aware module (UAM) to quantify the uncertainty of each object using cross-modal and illumination estimation, improving the detection benchmarks in multi-modal oriented detection tasks. 
TSFADet \cite{yuan2022translation} introduces translation-scale-rotation alignment (TSRA) to alleviate cross-modal misalignment in RGB-IR images by calibrating the feature maps of both modalities.
C2Former \cite{yuan2024c} designs an intermodality cross-attention (ICA) and adaptive feature sampling (AFS) to address the issues of inaccurate modality calibration and fusion, as well as the high computational cost of global attention.
CAGTDet \cite{yuan2024improving} enhances the model's regional feature fusion capability through the complementary fusion transformer (CFT).
CMAFF \cite{qingyun2022cross} combines common-mode and differential-mode attention to infer attention maps in parallel for adaptive feature enhancement, making the model more lightweight.
SuperYOLO \cite{zhang2023superyolo} utilizes auxiliary super-resolution learning \cite{li2025enhanced, li2024model, li2023x, li2023model} for multi-scale object detection, achieving good detection performance by detecting objects at high resolution. However, it is trained with horizontal box supervision, which limits precise localization of objects.
YOLOFIV \cite{wang2024yolofiv} achieves efficient object detection under complex illumination conditions through a designed dual-stream backbone network and enhanced rotation detection head, while maintaining high detection accuracy and computational speed.
E2E-MFD \cite{zhang2024e2e} achieves multi-scale feature extraction and target detection optimization from coarse to fine through the use of an object-region-pixel phylogenetic tree (ORPPT) and coarse-to-fine diffusion processing (CFDP).
DMM \cite{zhou2024dmm} is based on the Mamba architecture and addresses the challenges posed by inter-modal differences and intra-modal variations in target detection by introducing disparity-guided crossmodal fusion mamba (DCFM) and multi-scale target-aware attention (MTA), alongside target-prior aware (TPA).
DDCINet \cite{bao2025dual} addresses modality inconsistency and redundancy by introducing Dual-Dynamic Cross-Modal Interaction (DDCI) and Dynamic Feature Fusion (DFF), significantly enhancing performance across multiple multimodal remote sensing object detection datasets.

Among the latest methods, some algorithms are adaptations of conventional horizontal bounding box multimodal detection algorithms to rotated bounding box multimodal remote sensing detection tasks. However, existing rotated bounding box multimodal oriented detectors predominantly employ Transformer architectures or adopt three-branch feature extraction frameworks with three-branch classification-regression head designs. Such stacking of feature extraction backbones and detection heads leads to issues of excessive model parameter size and high computational complexity.

\subsection{Joint Label Training}
In multispectral fusion detection scenarios, annotation discrepancies often exist across modalities, and such inconsistency may lead to deviations in model optimization directions.
Sun et al.\cite{sun2022drone, yuan2024c, yuan2022translation} recognized this issue and therefore employed infrared spectral labels that are closer to real-world conditions to train models with multimodal inputs, potentially leading to suboptimal performance in the resulting multimodal models.
Yuan \cite{yuan2024improving} inserted labels from one modality into those of another modality, but this simple label stacking method ignores the modality specificity (spatial and semantic biases) of the labels.
In multi-task learning, models are required to simultaneously handle diverse tasks (e.g., detection and segmentation \cite{chen2019hybrid, wu2022yolop}, image super-resolution and detection \cite{zhang2023superyolo, hu2024beyond, wang2024degradation, huang2025crkd, deng2024polsar}), and thus commonly employ joint training strategies that dynamically adjust task-specific weights to optimize the model collectively.
Several studies \cite{kim2020cua,kendall2017uncertainties,sensoy2018evidential, tian2024weighted} also focus on general-purpose joint label training approaches, employing dynamically adjusted loss functions to integrate diverse data labels during training.

Inspired by these works, this paper introduces a joint label training approach into multispectral oriented detection tasks to mitigate model optimization bias induced by cross-modal annotation inconsistencies.

\section{Methodology}

This section presents the proposed two-stage oriented detection framework for multispectral RSI, as illustrated in Fig. \ref{MO_R-CNN}, which consists of three main components: the heterogeneous feature extraction network (HFEN), single modality supervision (SMS), and condition-based multimodal label fusion (CMLF).
In Section A, we will focus on introducing the heterogeneous characteristics of HFEN. Then, in Section B, we will describe the proposed SMS and discuss the loss function. Finally, the label fusion rules specified by CMLF will be presented in Section C.

\begin{figure}[!t]
	\centering
	\includegraphics[width=\columnwidth]{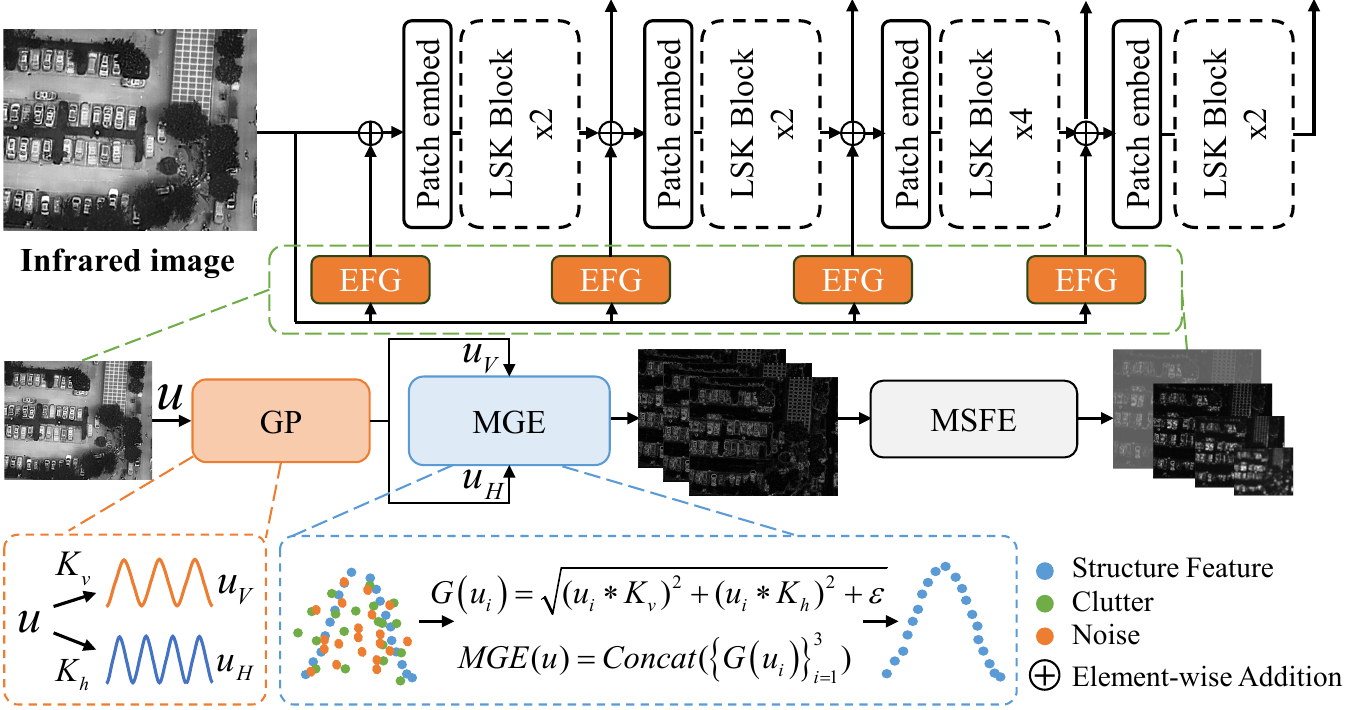}
	\caption{
		Overall framework of IRSI feature extraction branch. 
		This branch enhances the feature representation capability of LSK-S for object edges through the proposed EFG module.
		The EFG extracts the edge information of IRSI through GP and MGE, and then generates four edge feature representations at different scales via MSFE.
		Finally incorporates these multi-scale edge features into the LSK-S feature extraction process to enhance the backbone network's focus on edge information.
	}
	\label{IRSI_feature_extraction_branch}
\end{figure}

\subsection{Heterogeneous Feature Extraction Network}
The HFEN consists of three components: the HDS-LSK, SMFF, and RFA.
The HDS-LSK comprises two differentiated branches: the VRSI feature extraction branch and the IRSI feature extraction branch, dedicated to performing feature extraction operations on VRSI and IRSI respectively. The following sections elaborate on each component.

\subsubsection{VRSI feature extraction branch}
This branch exclusively employs LSK-S \cite{li2023large}, which operates in four progressive stages to extract multi-scale features across four distinct scales.
The features of each stage first pass through a Patch Embed module for feature embedding, adjusting the channel dimensions to \{64, 128, 320, 512\} respectively. 
Afterwards, they sequentially pass through \{2, 2, 4, 2\} LSK Blocks \cite{li2023large} and normalization layers to yield multi-scale features in the visible modality.

\subsubsection{IRSI feature extraction branch}
To fully leverage the edge contours of IRSI, we propose the edge feature guidance (EFG) to direct the LSK-S \cite{li2023large} towards enhanced focus on infrared object contour information, thereby introducing the IRSI feature extraction branch, as illustrated in Fig. \ref{IRSI_feature_extraction_branch}.
The EFG extracts multi-directional edge information of IRSI through the gradient processor (GP) and mixed gradient equation (MGE) and then generates four edge feature representations at different scales via the multi-scale feature encoder (MSFE).
Finally, it incorporates these multi-scale edge features into the feature extraction process of LSK-S \cite{li2023large} to enhance the focus of the backbone on edge information.
The input of MEG was generated from IRSI via GP. In the GP, two filters $K_v$ and $K_h$ were employed to process IRSI to extract vertical and horizontal edge information on each channel, respectively.
$K_v$ and $K_h$ are represented as follows:
\begin{equation}
	K_v = 	\left[ 
	\begin{array}{ccc}
		0 & -1 & 0\\
		0 & 0  & 0\\
		0 & 1  & 0
	\end{array} 
	\right];\quad
	K_h = 	\left[
	\begin{array}{ccc}
		0  & 0 & 0\\
		-1 & 0 & 1\\
		0  & 0 & 0
	\end{array}
	\right]
\end{equation}

Based on the above $K_v$ and $K_h$, the processing procedure of GP can be described as:
\begin{equation}
	u_v = Concat(\{ {u_i} * {K_v}\} _{i = 1}^3)
\end{equation}
\begin{equation}
	u_h = Concat(\{ {u_i} * {K_h}\} _{i = 1}^3)
\end{equation}
where $u_i$ represents the $i$-th channel of IRSI, and $Concat( \cdot )$ denotes the channel-wise concatenation operation.

The MGE integrates edge information from multiple directions of IRSI to generate a more robust edge representation, the process that can be formulated as:
\begin{equation}
	G({u_i}) = \sqrt {{{({u_i} * {K_v})}^2} + {{({u_i} * {K_h})}^2} + \varepsilon } ,\;i = 1,2,3
\end{equation}
\begin{equation}
	\begin{array}{l}
		x = MGE(u) = Concat(\left\{ {G({u_i})} \right\}_{i = 1}^3)
	\end{array}
\end{equation}
where $\varepsilon$ denotes a small non-zero positive value (set to $1\times10^6$) to prevent numerical instability during square root computation, $Concat( \cdot )$ represents the channel-wise concatenation operation, and $x \in \mathbb{R}^{3 \times H \times W}$ corresponds to the edge feature representation output by the MGE.
$x$ filtered out non-structural features such as clutter and noise information, serving as the input for MSFE.

\begin{figure}[!t]
	\centering
	\includegraphics[width=1.0\columnwidth]{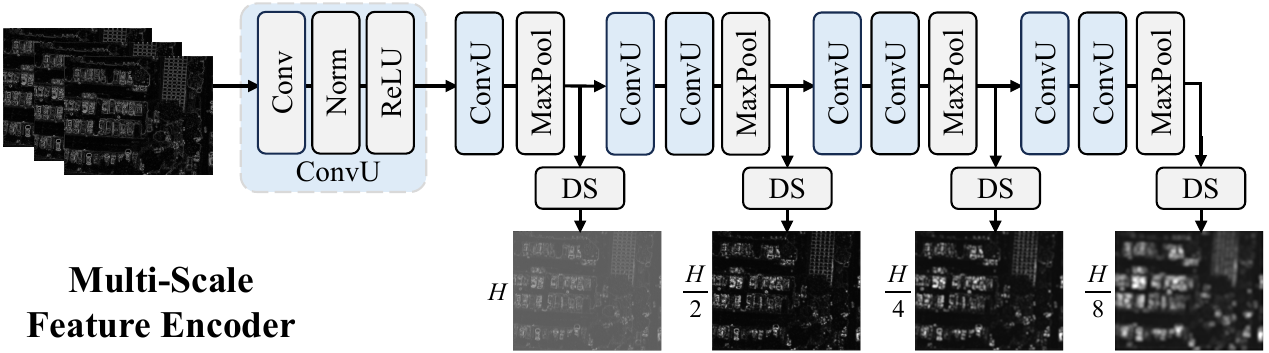}
	\caption{
		Overall framework of multi-scale feature encoder (MSFE). 
		The MSFE encodes the edge features obtained from MEG into multi-scale edge features that align with the backbone’s hierarchical features.
	}
	\label{Multi-Scale_Feature_Encoder}
\end{figure}

\begin{figure*}[!t]
	\centering
	\includegraphics[width=0.9\textwidth]{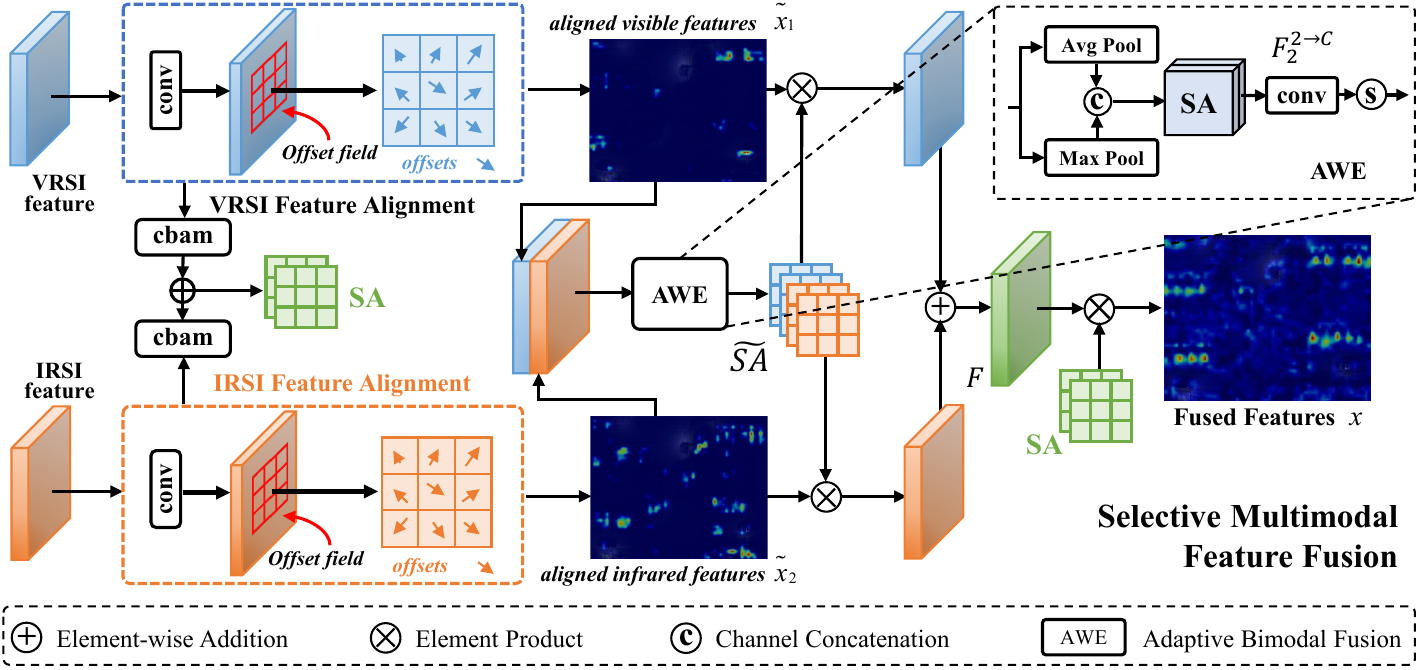}
	\caption{
		Overall framework of selective multimodal feature fusion. 
		The SMFF first predicts the offset field for deformable convolution through an offset generation network, then applies deformable convolution to achieve spatial alignment of bimodal features, and finally incorporates attention mechanisms to realize adaptive weighted fusion of features.
	}
	\label{Selective_Multimodal_Feature_Fusion}
\end{figure*}

The MSFE encodes the edge features obtained from the MGE into multi-scale edge features that align with the hierarchical features of the backbone network, as illustrated in Fig. \ref{Multi-Scale_Feature_Encoder}.
For ease of description, this paper denotes the integrated unit comprising convolution (Conv), normalization (Norm), and rectified linear unit (ReLU) as ConvU.
Each layer of the MSFE consists of two ConvU units and a MaxPool layer.
The first of the two ConvU units is responsible for channel alignment, while the second facilitates continuous feature learning. 
Furthermore, to prevent excessive information loss during feature downsampling by MaxPool layers, the MSFE adaptively adjusts feature scales by integrating both MaxPool and trilinear interpolation (Trilinear).
In the case of $4\times$ downsampling (DS), the encoder first applies a $2 times$ MaxPool operation, followed by a $2 times$ Trilinear interpolation to progressively reduce the feature scale.
This progressive downsampling strategy enables the multi-scale encoder to better balance performance and parameter efficiency.

\subsubsection{Selective Multimodal Feature Fusion}
In HDS-LSK, single-stream features propagate independently within each branch to avoid inter-modal interference during feature extraction. 
However, this design imposes stricter performance requirements on the subsequent feature fusion module.
To optimize multimodal data fusion capabilities, we propose the selective multimodal feature fusion (SMFF), with its architecture illustrated in Fig. \ref{Selective_Multimodal_Feature_Fusion}.
The SMFF aligns multi-scale features via deformable convolution and integrates an attention mechanism to achieve adaptive weighted feature fusion.
Specifically, given the feature maps ${x_1}$ and ${x_2}$ of VRSI and IRSI, the offset fields for deformable convolution are first predicted by an offset generation network:
\begin{equation}
	\Delta {p_1} = OffsetCon{v_1}({x_1})
\end{equation}
\begin{equation}
	\Delta {p_2} = OffsetCon{v_2}({x_2})
\end{equation}
where ${x_1} \in \mathbb{R}^{C \times H \times W}$, ${x_2} \in \mathbb{R}^{C \times H \times W}$ denote the feature maps of VRSI and IRSI, $OffsetCon{v_1}( \cdot )$ and $OffsetCon{v_2}( \cdot )$ represent the convolutional operations for predicting offset fields, and $\Delta {p_1}$ and $\Delta {p_1}$ correspond to the offset values required for feature alignment of ${x_1}$ and ${x_2}$, respectively.
Subsequently, deformable convolution is applied to spatially align the dual-modal features:
\begin{equation}
	{\widetilde x_1} = DeformConv({x_1},\Delta {p_1})\\
\end{equation}
\begin{equation}
	{\widetilde x_2} = DeformConv({x_2},\Delta {p_2})
\end{equation}
where $DeformConv( \cdot )$ denotes the deformable convolution operation, and ${\widetilde x_1}$ and ${\widetilde x_2}$ represent the aligned single-modal features.
To enhance the robustness of the fused feature representations, the cbam attention mechanism \cite{woo2018cbam} is applied to the aligned features ${\widetilde x_1}$, ${\widetilde x_2}$ to generate the fusion attention matrix:
\begin{equation}
	SA = cbam({\widetilde x_1}) \oplus cbam({\widetilde x_2})
\end{equation}
where $\oplus$ represents element wise addition operation.
On the other hand, the SMFF dynamically selects the most appropriate feature fusion strategy based on the characteristics of input data ${\widetilde x_1}$ or ${\widetilde x_2}$, performing adaptive fusion to enhance the model's adaptability and performance in complex application scenarios:
\begin{equation}
	\widetilde {SA} = \{ {\widetilde {SA}_1},{\widetilde {SA}_2}\}  = AWE(Concat({\widetilde x_1},{\widetilde x_2}))
\end{equation}
where $Concat( \cdot )$ denotes the channel-wise concatenation operation, ${\widetilde {SA}_1}$ and${\widetilde {SA}_2}$represent the adaptive fusion weight matrices, and $AWE( \cdot )$ corresponds to the adaptive weight extractor (AWE).
The AWE achieves adaptive weight extraction through average pooling and max pooling, that can be formulated as:
\begin{equation}
	M = Concat({\widetilde x_1},{\widetilde x_2})
\end{equation}
\begin{equation}
	\{ {\widetilde {SA}_1},{\widetilde {SA}_2}\} = AWE(M) = Con{v^{2 -  > N}}(A)
\end{equation}
\begin{equation}
	A = Concat\{ Avg(M),Max(M)\}
\end{equation}
where, $Con{v^{2->C}}( \cdot )$ denotes the dimensionality-adjusting convolution, $C$ represents the number of feature channels.
The final output of SMFF is jointly generated by ${\widetilde x_1}$, ${\widetilde x_2}$, $SA$, ${\widetilde {SA}_1}$, and ${\widetilde {SA}_2}$:
\begin{equation}
	x = \{ ({\widetilde x_1} \otimes {\widetilde {SA}_1}) \oplus ({\widetilde x_2} \otimes {\widetilde {SA}_2})\}  \otimes SA
\end{equation}

To more intuitively demonstrate the advantages of SMFF, this paper presents the fusion effects of SMFF in the ablation experiments section.
Subsequent experiments also show that SMFF significantly enhances the robustness of detection models in day-night scenarios.

\subsubsection{Residual Feature Augmentation}
The feature-fused feature pyramid generated by HDS-LSK and SMFF does not incorporate noise constraints, which makes the model vulnerable to background noise interference from both modalities.
In the FPN \cite{lin2017feature} of the baseline, the highest-level feature $M_4$ is propagated through the top-down pathway and gradually fused with the lower-level features $M_3$, $M_2$, and $M_1$.
$M_3$, $M_2$, and $M_1$ are dual-scale fused features with strong noise robustness. However, the highest-level feature $M_4$ contains only single-scale contextual information, making it highly susceptible to noise interference and resulting in poor feature representations.
To alleviate this issue, we introduce the RFA module \cite{guo2020augfpn}, which fuses higher-dimensional features at the original scale while optimizing $M_4$ to refine the final pyramid structure, thereby enhancing the model robustness.

\subsection{Single Modality Supervision}

The SMS consists of three components: visible modal supervision (VMS), infrared modal supervision (IMS), and detection head (DH). Compared to detection heads directly employing three independent branches for training and inference, SMS exhibits the following distinct characteristics:

\subsubsection{Shared Proposal Network in the First Stage}
Since RPN is a binary classification head for proposing obejct regions, we design VMS, IMS, and DH to share the same RPN proposal head. This architecture enables parameter sharing among them in the first stage, which not only reduces the parameter count but also enhances the proposal capability of RPN.

\subsubsection{Differentiated Feature Extraction Strategy in the Second Stage}
\begin{figure}[!t]
	\centering
	\includegraphics[width=\columnwidth]{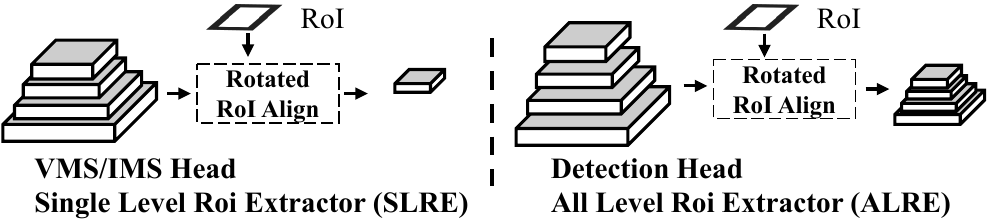}
	\caption{
		ALRE and SLRE. 
		The DH maps each RoI to the corresponding-level features on the fused feature pyramid $M_{fuse}$, employing SLRE. In contrast, VMS and IMS map each RoI to all feature layers of the single-spectrum pyramids$M_{vl}$ and $M_{ir}$, respectively, using ALRE.
	}
	\label{ALRE_and_SLRE}
\end{figure}

The inherent semantic gaps between scales in FPN \cite{lin2017feature} make it more suitable for scenarios with balanced distributions of large, medium, and small objects. Therefore, its direct application to small-object-dominated RSI detection may lead to suboptimal feature pyramids \cite{yang2023afpn}.
Most existing improvements focus solely on enhancing feature fusion mechanisms \cite{ghiasi2019fpn, liu2019learning, wang2019carafe, guo2020augfpn, zhang2020feature, ma2020dual, zhao2021graphfpn, huang2021fapn, xie2022latent, zhu2022improved, yang2023afpn}, which often increases model parameter counts or reduces inference speed.
This provides insights for the design of SMS: applying uniform supervision signals to multi-scale features before fusion to narrow their semantic gaps, thereby making the model more suitable for RSI tasks dominated by small-object instances.
As shown in Fig. \ref{ALRE_and_SLRE}, assuming the RoIs are proposals generated by the shared RPN among VMS, IMS, and DH. The DH maps each RoI to corresponding-level features on the fused feature pyramid $M_{fuse}$ using the single level roI extractor (SLRE), which helps enhance the multi-scale detection capability of model. Meanwhile, VMS and IMS map each RoI to all feature layers of the single-spectrum pyramids $M_{vl}$ and $M_{ir}$ respectively via the all level roI extractor (ALRE), enabling the second-stage classification and regression heads to share parameters across levels and learn more consistent semantic representations from heterogeneous spectral features.

\subsubsection{Discrepancies Between Training and Inference Strategies}
In existing methods, the three-branch detection heads are jointly trained and utilized during inference, which often increases parameter of model or reduces its inference speed. In contrast, our design enables DH to govern both training and inference processes, while VMS and IMS are only active during the training phase and removed at inference.
During model training, VMS and IMS compute losses for VRSI and IRSI labels respectively, enabling multimodal learning capabilities while constraining their corresponding unimodal feature pyramids.
To stabilize training, we employ $\lambda$ and $\eta$ to balance the unimodal supervision losses generated by VMS and IMS and the detection head loss from DH. The $Loss_{rcnn}$ of SMS is formulated as:
\begin{equation}
	Los{s_{rcnn}} = Los{s_{fuse}} + \lambda  \cdot Los{s_{vrsi}} + \eta  \cdot Los{s_{irsi}}
\end{equation}
where $\lambda$ and $\eta$ are set to 0.0625, $Loss_{fuse}$ denotes the regression loss of the second-phase DH detection head, and $Loss_{vrsi}$ and $Loss_{irsi}$ represent the losses from the VMS and IMS detection heads, respectively. Their mathematical formulations are as follows:
\begin{equation}
	\begin{aligned}
		Los{s_{fuse}} = & \sum\limits_{P = 2}^5 \Bigg[ L_{cls,{M_{fuse}}}({b_{fuse}},t_{fuse}^*) \\
		& + \beta [{t^*} > 0]{L_{loc,{M_{fuse}}}}({d_{fuse}},b_{fuse}^*) \Bigg]
	\end{aligned}
\end{equation}
\begin{equation}
	\begin{aligned}
		Los{s_{vrsi}} = & \sum\limits_{P = 2}^5 \Bigg[ L_{cls,{M_{vl}}}({b_{vl}},t_{vl}^*) \\
		& + \beta [{t^*} > 0]{L_{loc,{M_{vl}}}}({d_{vl}},b_{vl}^*) \Bigg]
	\end{aligned}
\end{equation}
\begin{equation}
	\begin{aligned}
		Los{s_{irsi}} = & \sum\limits_{P = 2}^5 \Bigg[ L_{cls,{M_{ir}}}({b_{ir}},t_{ir}^*) \\
		& + \beta [{t^*} > 0]{L_{loc,{M_{ir}}}}({d_{ir}},b_{ir}^*) \Bigg]
	\end{aligned}
\end{equation}
where ${L_{cls,{M_{vl}}}}( \cdot )$ are the objective functions attached to the unimodal feature pyramid ${M_{vl}}$, 
${L_{cls,{M_{ir}}}}( \cdot )$ and ${L_{loc,{M_{ir}}}}( \cdot )$ are the objective functions attached to the unimodal feature pyramid ${M_{ir}}$,
${L_{cls,{M_{fuse}}}}( \cdot )$ and ${L_{loc,{M_{fuse}}}}( \cdot )$ are the objective functions attached to the fused feature pyramid ${M_{fuse}}$,
$b$ and $d$ are the predicted values from the corresponding pyramid levels,
${t^*}$ and ${b^*}$ denote the ground-truth category labels and regression labels, respectively,
$\beta$ is used to balance the weights between classification and localization losses,
and $[{t^*} > 0]$ is defined as:
\begin{equation}
	[{t^*} > 0] = \left\{ \begin{array}{l}
		1,\;{t^*} > 0\\
		0,\;{t^*} = 0
	\end{array} \right.
\end{equation}

During the training phase, SMS employs joint training with single-spectrum labels and CMLF labels. In the inference phase, removing the VMS and IMS detection heads not only enhances the inference efficiency of model but also reduces its redundancy.

\begin{figure}[!t]
	\centering
	\includegraphics[width=1\columnwidth ]{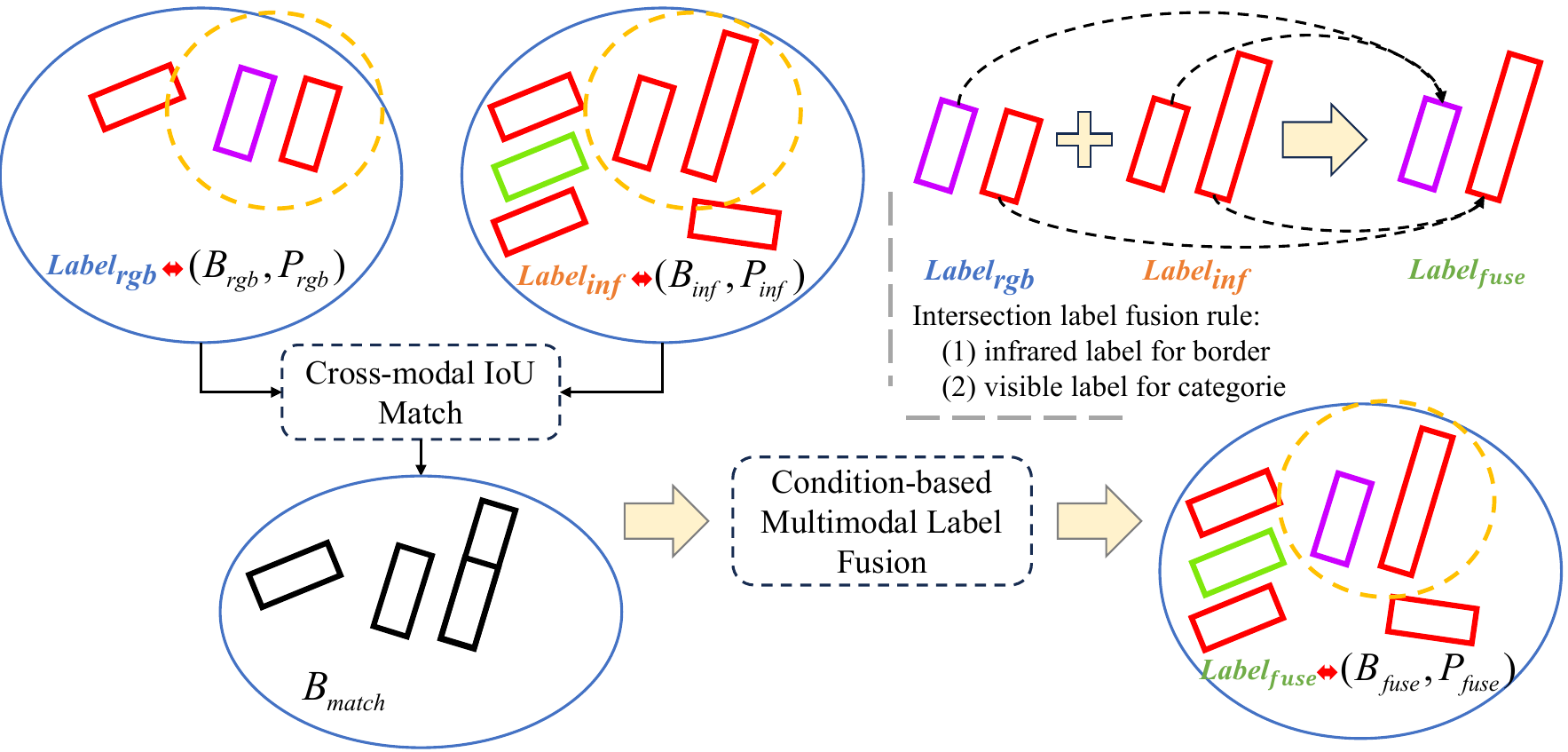}
	\caption{
		Overall framework of CMLF. 
		CMLF first calculates the cross-modal IoU (CMIoU) between bounding boxes from different modalities through CMIoU Matching. It then performs CMLF label fusion based on visible light labels, infrared labels, and the intersection set of bounding boxes. During this process, non-overlapping annotations are directly combined by taking the union, while overlapping annotations are fused according to the defined fusion rules.
	}
	\label{Condition-based_Multimodal_Label_Fusion}
\end{figure}

\subsection{Condition-based Multimodal Label Fusion}

To address annotation inconsistency issues in multispectral data sources for practical applications, we propose the condition-based multimodal label fusion (CMLF), as shown in Fig. \ref{Condition-based_Multimodal_Label_Fusion}.

CMLF calculates cross-modal IoU match (CMIoU Match) to obtain the CMIoU matrix between cross-modal bounding boxes, which can be described as:
\begin{equation}
	CMIoU_{}^{(i,j)} = \frac{{area(B_{_{rgb}}^{(i)} \cap B_{_{inf}}^{(j)})}}{{area(B_{_{rgb}}^{(i)} \cup B_{_{inf}}^{(j)})}}\;
\end{equation}
where $0 \le i \le m$, $0 \le j \le n$, with $m$ and $n$ denoting the number of bounding boxes in visible and infrared annotations respectively,
${B_{rgb}} \in \mathbb{R}{^m}$ and ${B_{inf}} \in \mathbb{R}{^n}$ represent the sets of visible and infrared annotated bounding boxes, 
while $CMIoU_{}^{(i,j)}$ indicates the rotational IoU value between the $i$-th bounding box in visible labels and the $j$-th bounding box in infrared labels.

Then, we define the bounding box set ${B_m}$ in the CMIoU matrix where CMIoU exceeds a specified threshold ($\tau $ =0.7. we consider two annotated boxes with over 70\% overlap area as representing the same target across modalities) as:
\begin{equation}
	{B_m} = \{ (i,j)|CMIoU_{}^{(i,j)} > \tau \}
\end{equation}

Finally, the CMLF performs label fusion based on labels of VRSI, labels of IRSI, and the bounding box intersection set ${B_m}$, where non-overlapping annotations are directly combined by taking their union, while overlapping annotations are fused according to the fusion rules specified by the following equation:
\begin{equation}
	\begin{array}{l}
		{B_{fuse}} = CMLF({B_{rgb}},{P_{rgb}},{B_{inf}},{P_{inf}},{B_{match}})\\
		= \{ b_{inf}^{(j)}|(i,j) \in {B_{match}}\}  \cup \{ b_{rgb}^{(i)}|\forall j,(i,j) \notin {B_{match}}\} \\
		\;\;\;\;	\cup \{ b_{inf}^{(j)}|\forall i,(i,j) \notin {B_{match}}\} 
	\end{array}
\end{equation}
\begin{equation}
	\begin{array}{l}
		{P_{fuse}} = CMLF({B_{rgb}},{P_{rgb}},{B_{inf}},{P_{inf}},{B_{match}})\\
		= \{ p_{rgb}^{(i)}|(i,j) \in {B_{match}}\}  \cup \{ p_{rgb}^{(i)}|\forall j,(i,j) \notin {B_{match}}\} \\
		\;\;\;\;   \cup \{ p_{inf}^{(j)}|\forall i,(i,j) \notin {B_{match}}\} 
	\end{array}
\end{equation}
where ${B_{rgb}} \in \mathbb{R}{^m}$ and ${P_{rgb}} \in \mathbb{R}{^m}$ denote the visible bounding box set and category set, respectively.
${B_{inf}} \in \mathbb{R}{^n}$, ${P_{inf}} \in \mathbb{R}{^n}$ represent the infrared bounding box set and category set, respectively. 
${B_{match}}$ indicates the intersection bounding box set, while 
${B_{fuse}}$ and ${P_{fuse}}$ correspond to the bounding box set and category set of the CMLF labels, respectively.

\begin{figure}[!t]
	\centering
	\includegraphics[width=\columnwidth ]{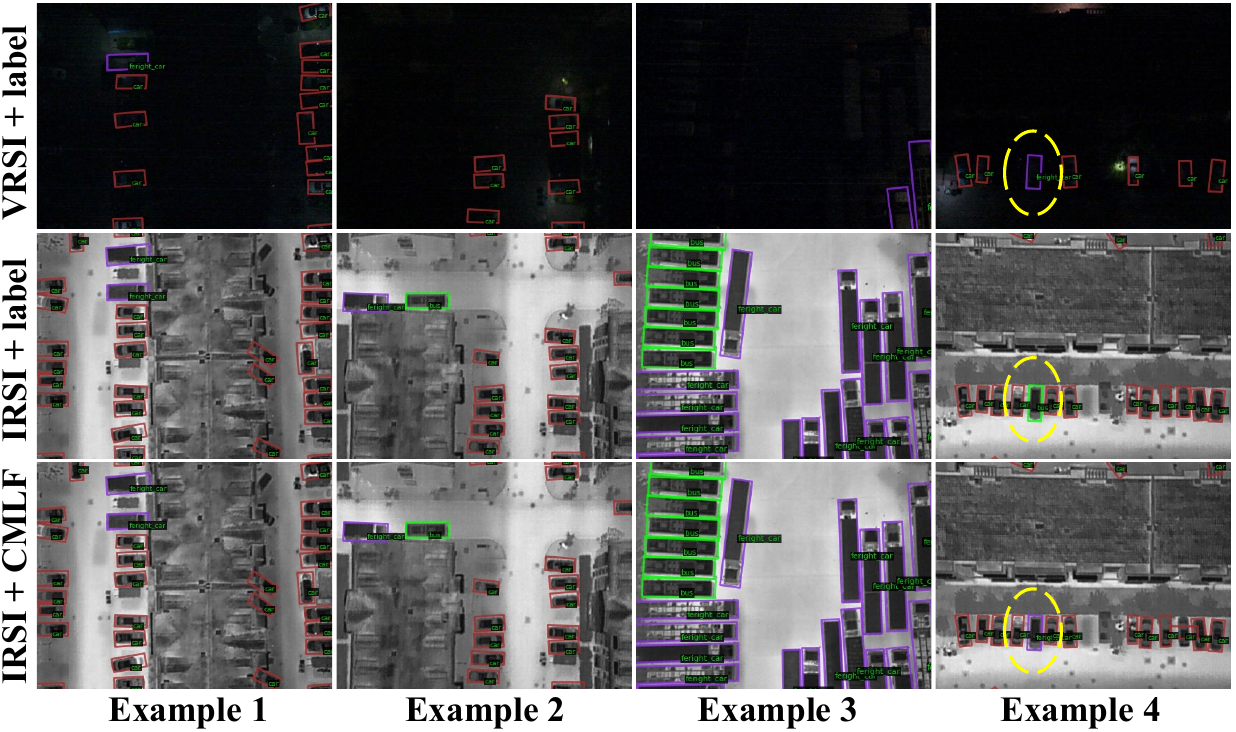}
	\caption{
		Label comparison diagram. 
		In the first three cases, CMLF labels are formed by having IRSI supplement both bounding boxes and category information of VRSI. 
		In the fourth case, while IRSI supplements the VRSI bounding boxes, VRSI simultaneously corrects the category labels of IRSI, as demonstrated by the target in the yellow region.
	}
	\label{Label_Comparison_Diagram}
\end{figure}

Fig. \ref{Label_Comparison_Diagram} illustrates the comparison between VRSI, IRSI, and CMLF fused labels.
In the first three cases, CMLF labels are formed by supplementing VRSI's bounding boxes and category information with IRSI annotations.
In the fourth case, while IRSI enhances VRSI's bounding boxes, VRSI simultaneously corrects IRSI's category labels (as indicated by the target in the yellow region).
All CMLF labels exhibit optimization compared to single-spectral labels, providing a reliable foundation for subsequent model training.

\section{Experiments}
In this section, we first introduce the datasets and evaluation metrics used in our experiments. 
We then provide comparisons with several state-of-the-art methods. 
Next, we conduct ablation studies and perform a detailed analysis of the effectiveness of our methods. 
Finally, we analyze and compare the computational cost of our method.

\subsection{Experiment Setup}

\subsubsection{Datasets}
DroneVehicle \cite{sun2022drone} is a largescale drone-based infrared-visible oriented object detection dataset.
It contains 28,439 pairs of RGB-IR images, covering various scenarios such as urban roads, residential areas, parking lots, and different times from day to night.
The dataset provides independently annotated oriented bounding box labels for infrared and visible modalities, covering five categories: car, bus, truck, van, and freight car.

VEDAI \cite{razakarivony2016vehicle} is designed for vehicle detection in high-resolution aerial images.
It includes diverse urban and rural scenes with oriented annotations for various vehicle types such as cars, trucks, and vans, along with a few other objects like airplanes and ships, totaling 9 categories.
The dataset consists of 1,246 pairs of RGB and infrared images with resolutions of $1024 \times 1024$ and $512 \times 512$ pixels.
In our experiments, we use the $512 \times 512$ version and follow the ten-fold cross-validation protocol as described in \cite{razakarivony2016vehicle} to train and test our model.

OGSOD \cite{wang2023category} is a recently released VRSI–SAR paired dataset for multimodal remote sensing object detection.
It consists of a training set of 14 665 image pairs and a test set of 3666 image pairs, containing over 48 000 instance annotations in total.
The image size for all images is $256 \times 256$, and in the experiment, the images were resized to $512 \times 512$.
Three categories, including bridges, storage tanks, and harbors, are annotated, and the rotated bounding box annotation format is used to evaluate our proposed methods.

\begin{table*}[!t]
	\renewcommand{\arraystretch}{0.85}
	\footnotesize
	\centering
	\caption{
		Comparison of Different Methods on DroneVehicle Dataset. 
		The methods marked with `$^{\dagger}$' denote our reimplementation of other algorithms, whereas those without such annotations correspond to the original data presented in the paper.
	}
	\label{Comparison_of_Different_Methods_on_DroneVehicle_Dataset}
	\begin{tabular*}{\textwidth}{@{\extracolsep{\fill}}c|l|c|c|c|ccccc|cc}
		\toprule
		\multirow{2}{*}{\textbf{Modality}} & \multirow{2}{*}{\textbf{Method}}  & \multirow{2}{*}{\textbf{Basic Detector}} & \multirow{2}{*}{\textbf{Label}} & \multirow{2}{*}{\textbf{Epoch}} & \multicolumn{5}{c}{\textbf{All categories in DroneVehicle}} & \multirow{2}{*}{\textbf{Params}}	& \multirow{2}{*}{\textbf{mAP}}  \\
		
		\cmidrule{6-10} 
		& & & & & \textbf{CA} & \textbf{TR} & \textbf{BU} & \textbf{VA} & \textbf{FC} & &  \\
		
		\midrule
		\multirow{8}{*}{VRSI}      		
		& R${}^3$Det$^{\dagger}$\cite{yang2021r3det} 			& - 			 & VRSI & 12 & 85.2 & 51.1 & 82.8 & 29.4 & 26.7 & 36.21 & 55.06 \\
		& Oriented RepPoints$^{\dagger}$\cite{li2022oriented}  	& -				 & VRSI & 12 & 88.6 & 60.1 & 85.5 & 41.0 & 39.5 & 36.60 & 62.95 \\
		& Gliding Vertex$^{\dagger}$\cite{xu2020gliding} 	 	& -				 & VRSI & 12 & 88.7 & 61.4 & 86.3 & 42.6 & 41.5 & 41.13 & 64.10 \\
		& S${}^2$ANet$^{\dagger}$\cite{han2021align} 			& -				 & VRSI & 12 & 88.6 & 62.3 & 87.0 & 42.7 & 42.7 & 38.58 & 64.63 \\
		& Oriented R-CNN$^{\dagger}$\cite{xie2021oriented} 	 	& -				 & VRSI & 12 & 88.8 & 64.5 & 87.9 & 43.3 & 41.7 & 41.13 & 65.23 \\
		& RoITransformer$^{\dagger}$\cite{ding2019learning} 	& -				 & VRSI & 12 & 89.0 & 65.8 & 88.6 & 45.7 & 44.3 & 55.06 & 66.66 \\
		& PKINet-S$^{\dagger}$\cite{pu2023adaptive} 			& Oriented R-CNN & VRSI & 12 & 89.2 & 68.7 & 88.9 & 48.0 & 47.5 & 30.85 & 68.45 \\
		& LSKNet-S$^{\dagger}$\cite{li2023large} 				& Oriented R-CNN & VRSI & 12 & 89.3 & 70.8 & 89.2 & 55.4 & 52.5 & 30.97 & 71.46 \\
		
		\midrule
		\multirow{8}{*}{IRSI}         
		& R${}^3$Det$^{\dagger}$\cite{yang2021r3det} 		 	& -				 & IRSI & 12 & 88.9 & 54.4 & 86.1 & 32.4 & 41.9 & 36.21 & 60.74 \\
		& Oriented RepPoints$^{\dagger}$\cite{li2022oriented}   & -				 & IRSI & 12 & 90.0 & 60.1 & 86.7 & 37.1 & 48.2 & 36.60 & 64.43 \\
		& Gliding Vertex$^{\dagger}$\cite{xu2020gliding} 	 	& -				 & IRSI & 12 & 90.0 & 61.1 & 87.7 & 41.7 & 46.9 & 41.13 & 65.50 \\
		& S${}^2$ANet$^{\dagger}$\cite{han2021align} 			& -				 & IRSI & 12 & 90.0 & 62.6 & 88.3 & 42.8 & 50.1 & 38.58 & 66.76 \\
		& Oriented R-CNN$^{\dagger}$\cite{xie2021oriented}  	& -				 & IRSI & 12 & 89.8 & 57.4 & 89.3 & 45.4 & 53.1 & 41.13 & 67.03 \\
		& RoITransformer$^{\dagger}$\cite{ding2019learning} 	& -				 & IRSI & 12 & 90.1 & 64.0 & 88.8 & 42.5 & 51.3 & 55.06 & 67.35 \\
		& PKINet-S$^{\dagger}$\cite{pu2023adaptive}				& Oriented R-CNN & IRSI & 12 & 90.2 & 72.6 & 88.7 & 51.3 & 56.3 & 30.85 & 71.85 \\
		& LSKNet-S$^{\dagger}$\cite{li2023large}				& Oriented R-CNN & IRSI & 12 & 90.2 & 75.1 & 89.2 & 53.7 & 59.8 & 30.97 & 73.62 \\
		
		\midrule
		\multirow{22}{*}{VRSI + IRSI}
		& UA-CMDet\cite{sun2022drone} 			 			 	& RoITransformer & IRSI & 12 & 87.5 & 60.7 & 87.1 & 38.0 & 46.8 & 138.69 & 64.01 \\
		& YOLOFIV\cite{wang2024yolofiv}    						& YOLOv5s 	 	 & -    & 300& 95.9 & 64.2 & 91.6 & 37.3 & 34.6 & 10.61  & 64.71 \\
		& Halfway Fusion\cite{wagner2016multispectral} 			& Faster R-CNN   & -	& -  & 89.9 & 60.3 & 89.0 & 46.3 & 55.5 & -	     & 68.19 \\
		& CIAN\cite{zhang2019cross} 					 		& -              & -    & -  & 90.0 & 62.5 & 88.9 & 49.6 & 66.2 & -	     & 70.23 \\
		& AR-CNN\cite{zhang2021weakly} 				 			& Faster R-CNN   & -    & -  & 90.1 & 64.8 & 89.4 & 51.5 & 62.1 & -	     & 71.58 \\
		& LF-MDet (HBB)\cite{sun2024low}   						& DINO 	 		 & -    & 18 & 82.2 & 73.6 & 86.6 & 57.0 & 59.6 & 38.7   & 71.80 \\
		& MBNet\cite{zhou2020improving} 					 	& -				 & -    & -  & 90.1 & 64.4 & 88.8 & 53.6 & 62.4 & -	     & 71.90 \\
		& TSFADet\cite{yuan2022translation} 				 	& Oriented R-CNN & IRSI & -  & 89.9 & 67.9 & 89.8 & 54.0 & 63.7 & 104.70 & 73.06 \\
		& Cascade-TSFADet\cite{yuan2022translation} 		 	& Cascade R-CNN  & IRSI & 20 & 90.0 & 69.2 & 89.7 & 55.2 & 65.5 & -	     & 73.90 \\
		& C2Former\cite{yuan2024c} 					 			& S$^2$ANet      & -    & 24 & 90.2 & 68.3 & 89.8 & 58.5 & 64.4 & 100.80 & 74.20 \\
		& GAGTDet\cite{yuan2024improving} 					 	& Oriented R-CNN & -    & 20 & 90.8 & 69.7 & 90.5 & 55.6 & 66.3 & -	     & 74.57 \\
		& DDCINet-S$^2$ANet\cite{bao2025dual}					& S$^2$ANet	     & -    & 12 & 90.4 & 75.5 & 89.7 & 60.9 & 62.0 & 121.71 & 75.70  \\
		& DMM\cite{zhou2024dmm}									& Faster R-CNN   & IRSI & 12 & 90.4 & 77.8 & 88.7 & 63.0 & 66.0 & 87.97	 & 77.20  \\
		& E2E-MFD\cite{zhang2024e2e}							& -			     & -    & 50 & 90.3 & 79.3 & 89.8 & 63.1 & 64.6 & -   	 & 77.40  \\
		& DDCINet-RoI-Trans\cite{bao2025dual}					& RoITransformer & -    & 12 & 90.5 & 78.6 & 90.0 & 64.9 & 64.9 & -      & 77.50  \\
		& MO R-CNN$^{\dagger}$ (ours)   						& Oriented R-CNN & VRSI & 36 & 90.0 & 80.6 & 89.8 & 66.8 & 64.6 & 54.90  & \textbf{78.36} \\
		& MO R-CNN$^{\dagger}$ (ours)    		 				& Oriented R-CNN & IRSI & 36 & 90.4 & 79.7 & 89.8 & 65.8 & 64.5 & 54.90  & \textbf{78.03} \\
		& MO R-CNN$^{\dagger}$ (ours)   						& Oriented R-CNN & CMLF & 36 & 89.6 & 78.7 & 88.3 & 64.1 & 63.6 & 54.90  & \textbf{76.84} \\
		\bottomrule
	\end{tabular*}
\end{table*}

\subsubsection{Implementation Details}
We use MMRotate \cite{zhou2022mmrotate} as the basic framework and implement experiments on 1 NVIDIA RTX3090 GPU.
The backbone networks utilized in this paper are pre-trained on ImageNet \cite{deng2009imagenet}.
In the training phase, we employ AdamW \cite{loshchilov2017decoupled} as our optimizer, setting the learning rate, weight decay, and momentum parameters to 0.01, 0.0001, and 0.9, respectively.
We adopted our own implemented data augmentation methods: TwoStreamRResize, TwoStreamRRandomFlip, and TwoStreamPolyRandomRotate.
The model following the $1 \times $ training strategy is trained for 12 epochs, with the learning rate being reduced by a factor of 10 in the 8th and 11th epochs. As for the model associated with the $3 \times$ training strategy, it is trained for 36 epochs, with a 10-fold reduction of the learning rate in the 27th and 33rd epochs.
Each mini-batch consists of 2 images processed on a single GPU.
The other parameters strictly follow the default configuration of MMRotate \cite{zhou2022mmrotate}.
The evaluation metrics selected were $A{P_{50}}$ and $mAP_{50}$.
When training DroneVehicle \cite{sun2022drone}, The model is trained on the training set, validated on the validation set, and tested on the test set.
Given the limited scale of the VEDAI \cite{razakarivony2016vehicle}, we adopted transfer learning. Specifically, we first pre-train the model on DroneVehicle \cite{sun2022drone} for 36 epochs, followed by fine-tuning on the VEDAI \cite{razakarivony2016vehicle}.

\begin{table*}[!t]
	\renewcommand{\arraystretch}{0.90}
	\footnotesize
	\centering
	\caption{
		Performance of Visual Backbone Based on Transformer or Mamba in MO R-CNN
		Compared to the Oriented R-CNN framework, Swin-S and VMamba-S demonstrated superior performance in MO R-CNN.
	}
	\label{Performance_of_Visual_Backbone_Based_on_Transformer_or_Mamba_in_MO_R-CNN}
	\begin{tabular*}{\textwidth}{@{\extracolsep{\fill}}c|l|c|ccccc|cl}
		\toprule
		\multirow{2}{*}{\textbf{Modality}} & \multirow{2}{*}{\textbf{Method}} & \multirow{2}{*}{\textbf{Label}}& \multicolumn{5}{c}{\textbf{All categories in DroneVehicle}} & \multirow{2}{*}{\textbf{Params}}	& \multirow{2}{*}{\textbf{mAP@0.5}}  \\
		
		\cmidrule{4-8} 
		& & & \textbf{CA} & \textbf{TR} & \textbf{BU} & \textbf{VA} & \textbf{FC} & &  \\
		\midrule
		\multirow{11}{*}{VRSI + IRSI}
		& DS-Swin-S$^{\dagger}$\cite{liu2021swin} + Oriented R-CNN		& IRSI & 90.4 & 78.7 & 89.6 & 58.1 & 61.7 & 116.48  & 75.97 \\
		& DS-Swin-S$^{\dagger}$\cite{liu2021swin} + MO R-CNN (ours)   	& IRSI & 90.4 & 78.8 & 89.8 & 59.9 & 61.6 & 131.65  & \textbf{76.09 (+0.12)} \\
		
		\cmidrule{2-10}
		& DS-Swin-S$^{\dagger}$\cite{liu2021swin} + Oriented R-CNN	  	& CMLF & 89.8 & 76.8 & 88.9 & 58.4 & 61.1 & 116.48  & 74.99 \\	
		& DS-Swin-S$^{\dagger}$\cite{liu2021swin} + MO R-CNN (ours)    	& CMLF & 89.6 & 77.4 & 88.0 & 60.0 & 60.4 & 131.65  & \textbf{75.10 (+0.11)} \\
		\cmidrule{2-10}
		& DS-VMamba-S$^{\dagger}$\cite{liu2024vmamba} + Oriented R-CNN	& IRSI & 90.4 & 78.9 & 89.5 & 58.8 & 63.6 & 117.56  & 76.23 \\
		& DS-VMamba-S$^{\dagger}$\cite{liu2024vmamba} + MO R-CNN (ours)	& IRSI & 90.4 & 80.1 & 89.8 & 63.8 & 64.0 & 132.73  & \textbf{77.62 (+1.39)} \\
		\cmidrule{2-10}
		& DS-VMamba-S$^{\dagger}$\cite{liu2024vmamba} + Oriented R-CNN	& CMLF & 89.9 & 77.2 & 88.5 & 61.3 & 61.7 & 117.56  & 75.71 \\						
		& DS-VMamba-S$^{\dagger}$\cite{liu2024vmamba} + MO R-CNN (ours)	& CMLF & 89.6 & 79.1 & 88.2 & 64.6 & 63.0 & 132.73  & \textbf{76.88 (+1.17)} \\	
		\bottomrule
	\end{tabular*}
\end{table*}

\begin{figure*}[!t]
	\centering
	\includegraphics[width=0.9\textwidth]{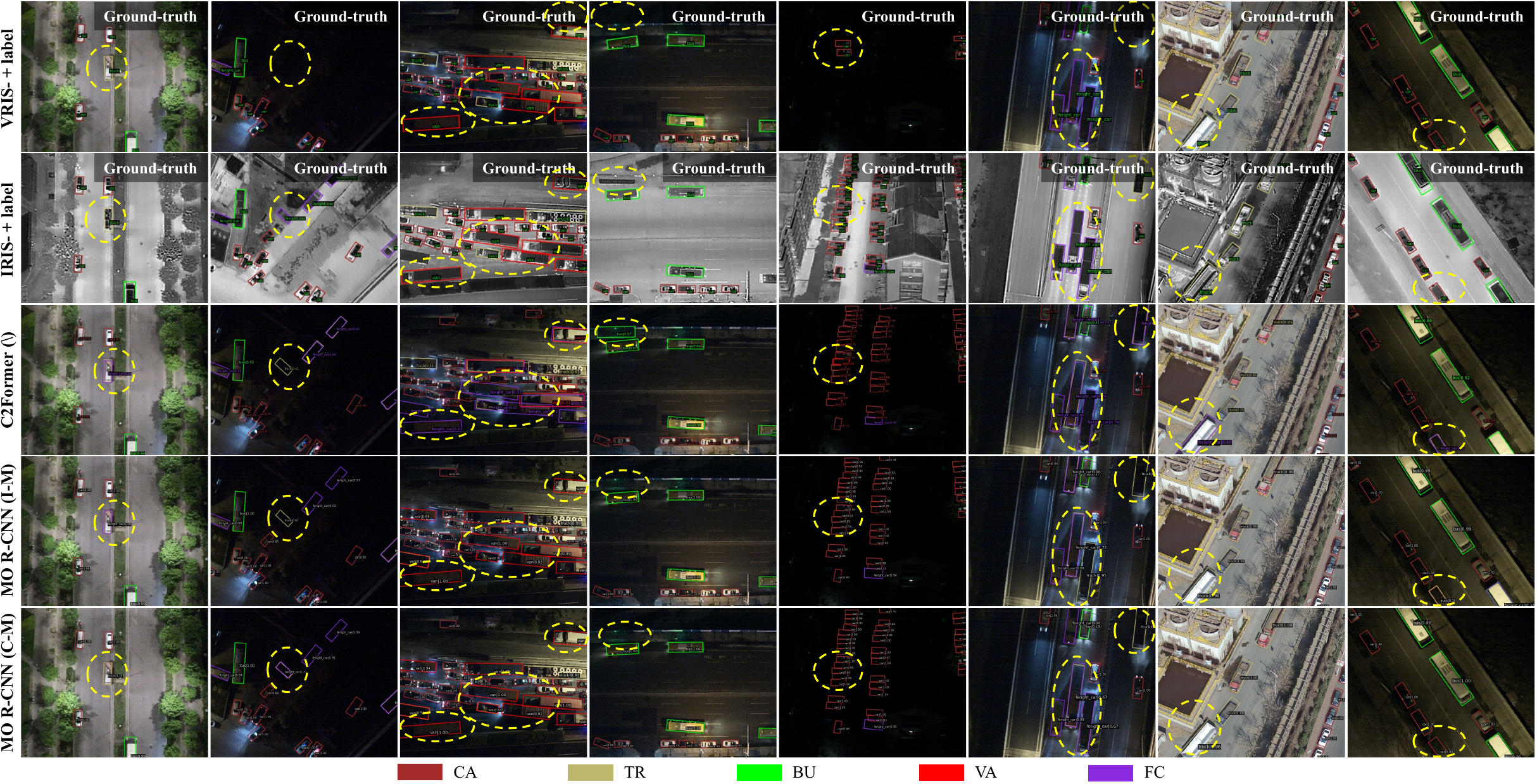}
	\caption{
		Visualization of detection results of recent methods and MO R-CNN on DroneVehicle. 
		From left to right, represent VRSI labels, IRSI labels, results of C2Former, results of IRSI label trained MO R-CNN(I-M), and results of CMLF label trained MO R-CNN (C-M), respectively.
	}
	\label{Visualization_of_Detection_Results_of_Recent_Methods_and_MO_R-CNN_on_DroneVehicle}
\end{figure*}

\begin{table*}[!t]
	\renewcommand{\arraystretch}{1.1}
	\footnotesize
	\centering
	\caption{
		Comparison of Different Methods on VEDAI Dataset. 
		Because the labels of the two modalities in the VEDAI dataset are exactly the same, CMLF labels were not used for training.
		The methods marked with `$^{\ddagger}$' represent our reproduction using transfer learning of algorithms proposed by other researchers, while the unmarked methods represent the original data from the original papers.
	}
	\label{Comparison_of_Different_Methods_on_VEDAI_Dataset}
	\begin{tabular*}{\textwidth}{@{\extracolsep{\fill}}c|l|c|c|c@{\hspace{5pt}}c@{\hspace{5pt}}c@{\hspace{5pt}}c@{\hspace{5pt}}c@{\hspace{5pt}}c@{\hspace{5pt}}c@{\hspace{5pt}}c@{\hspace{5pt}}c|c}
		\toprule
		\multirow{2}{*}{\textbf{Modality}} & \multirow{2}{*}{\textbf{Method}} & \multirow{2}{*}{\textbf{Basic Detector}} & \multirow{2}{*}{\textbf{Epoch}} & \multicolumn{9}{c}{\textbf{All categories in VEDAI}}	& \multirow{2}{*}{\textbf{mAP}}  \\
		
		\cmidrule{5-13} 
		& & & & \textbf{CA} & \textbf{TR} & \textbf{TRA} & \textbf{CC} & \textbf{VA} & \textbf{PI} & \textbf{BO} & \textbf{PL} & \textbf{OT} &   \\
		
		\midrule
		\multirow{8}{*}{VRSI}      		
		& R${}^3$Det$^{\ddagger}$\cite{yang2021r3det}				& -				 & 12  & 69.1 & 39.9 & 38.6 & 63.9 & 33.8 & 56.2 & 13.7 & 19.8  & 29.8 & 40.5 \\
		& Oriented RepPoints$^{\ddagger}$\cite{li2022oriented}  	& -				 & 12  & 76.2 & 66.8 & 58.3 & 77.4 & 51.1 & 70.8 & 31.7 & 74.3  & 45.5 & 61.3 \\
		& Gliding Vertex$^{\ddagger}$\cite{xu2020gliding} 	 		& -				 & 12  & 77.4 & 70.4 & 52.2 & 76.9 & 62.9 & 68.6 & 24.9 & 69.0  & 46.0 & 60.9 \\
		& S${}^2$ANet$^{\ddagger}$\cite{han2021align} 				& -				 & 12  & 77.3 & 63.8 & 59.1 & 75.5 & 52.9 & 71.0 & 21.0 & 75.4  & 45.3 & 60.1 \\
		& Oriented R-CNN$^{\ddagger}$\cite{xie2021oriented} 	 	& -				 & 12  & 82.1 & 73.3 & 58.1 & 80.3 & 70.4 & 75.4 & 49.4 & 78.1  & 49.8 & 68.5 \\
		& RoITransformer$^{\ddagger}$\cite{ding2019learning} 		& -				 & 12  & 81.8 & 71.1 & 66.6 & 79.6 & 70.6 & 77.5 & 47.9 & 52.1  & 71.3 & 71.3 \\
		& PKINet-S$^{\ddagger}$\cite{pu2023adaptive} 				& Oriented R-CNN & 12  & 84.9 & 76.7 & 70.9 & 80.2 & 72.9 & 81.8 & 66.5 & 79.2  & 54.9 & 74.2 \\
		& LSKNet-S$^{\ddagger}$\cite{li2023large} 					& Oriented R-CNN & 12  & 85.1 & 81.5 & 69.0 & 80.8 & 72.1 & 81.4 & 64.6 & 90.4  & 55.8 & \textbf{75.6} \\
		
		\midrule
		\multirow{8}{*}{IRSI}         
		& R${}^3$Det$^{\ddagger}$\cite{yang2021r3det} 		 		& -				 & 12  & 65.7 & 37.1 & 18.2 & 59.4 & 29.4 & 57.1 & 11.2 & 28.5  & 16.4 & 35.9 \\
		& Oriented RepPoints$^{\ddagger}$\cite{li2022oriented}  	& -				 & 12  & 73.6 & 53.0 & 35.3 & 68.9 & 45.0 & 66.2 & 17.2 & 71.3  & 30.0 & 51.2 \\
		& Gliding Vertex$^{\ddagger}$\cite{xu2020gliding} 	 		& -				 & 12  & 72.9 & 62.3 & 29.6 & 70.1 & 60.5 & 66.6 & 21.2 & 78.2  & 26.6 & 54.2 \\
		& S${}^2$ANet$^{\ddagger}$\cite{han2021align} 				& -				 & 12  & 75.4 & 53.9 & 32.4 & 68.6 & 43.4 & 67.8 & 14.2 & 49.5  & 37.1 & 47.8 \\
		& Oriented R-CNN$^{\ddagger}$\cite{xie2021oriented}  		& -				 & 12  & 78.2 & 70.6 & 42.9 & 75.6 & 68.6 & 72.8 & 44.2 & 79.9  & 35.9 & 63.2 \\
		& RoITransformer$^{\ddagger}$\cite{ding2019learning} 		& -				 & 12  & 79.1 & 68.6 & 43.7 & 75.5 & 66.0 & 73.0 & 38.9 & 89.5  & 33.3 & 63.1 \\
		& PKINet-S$^{\ddagger}$\cite{pu2023adaptive}				& Oriented R-CNN & 12  & 82.2 & 74.7 & 50.6 & 75.2 & 70.6 & 77.6 & 57.1 & 88.0  & 34.8 & 67.9 \\
		& LSKNet-S$^{\ddagger}$\cite{li2023large}					& Oriented R-CNN & 12  & 82.5 & 77.0 & 52.8 & 79.1 & 74.4 & 78.1 & 58.1 & 90.4  & 40.0 & \textbf{70.2} \\
		
		\midrule
		\multirow{9}{*}{VRSI + IRSI}  	
		& C2Former\cite{yuan2024c}    								& S$^2$ANet      & 12  & 76.7 & 52.0 & 59.8 & 63.2 & 48.0 & 68.7 & 43.3 & 47.0  & 41.9 & 55.6 \\
		& DMM \cite{zhou2024dmm}   									& S$^2$ANet      & 12  & 77.9 & 59.3 & 68.1 & 70.8 & 57.4 & 75.8 & 61.2 & 77.5  & 43.5 & 65.7 \\
		& CMAFF\cite{qingyun2022cross}  							& Oriented R-CNN & 12  & 81.7 & 58.8 & 68.7 & 78.4 & 68.5 & 76.3 & 66.0 & 72.7  & 51.5 & 69.2 \\
		& DMM\cite{zhou2024dmm}   									& Oriented R-CNN & 12  & 84.2 & 65.7 & 72.3 & 79.0 & 72.5 & 78.8 & 72.3 & 93.6  & 56.2 & 75.0 \\
		& Super YOLO(HBB)\cite{zhang2023superyolo}    				& YOLOv5s		 & 300 & 91.1 & 70.2 & 80.4 & 79.3 & 76.5 & 85.7 & 60.2 & -     & 57.3 & 75.1 \\
		& DDCINet\cite{bao2025dual}    								& Oriented R-CNN & 12  & -    & -    & -    & -    & -    & -    & -    & -     & -    & 77.0 \\
		& YOLOFIV\cite{wang2024yolofiv}    							& YOLOv5s        & 300 & 93.9 & 80.1 & 82.2 & 82.1 & 79.3 & 87.4 & 75.5 & -     & 60.8 & 80.2 \\
		& LF-MDet (HBB)\cite{sun2024low}    						& DINO           & 18  & 92.1 & 71.6 & 89.5 & 85.9 & 84.1 & 87.3 & 68.8 & -     & 67.1 & 80.8 \\
		& EMCFormer (HBB)\cite{wang2025emcformer}  						& YOLOv5s 	 	 & 400 & -    & -	 & -	& -	   & -	  & -    & -    & -     & -    & 84.7 \\
		& MO R-CNN$^{\ddagger}$ (ours)    							& Oriented R-CNN & 12  & 89.6 & 90.6 & 87.8 & 90.1 & 87.4 & 88.9 & 81.8 & 99.0  & 76.1 & 87.9 \\
		& MO R-CNN$^{\ddagger}$ (ours)    							& Oriented R-CNN & 36  & 90.8 & 99.7 & 90.9 & 98.9 & 95.8 & 90.8 & 90.7 & 100.0 & 90.5 & \textbf{94.2} \\
		
		\bottomrule
	\end{tabular*}
\end{table*}

\subsection{Comparisons of Multimodal Detection Models}
This section compares the proposed algorithm with other multimodal object detectors on DroneVehicle \cite{sun2022drone}, VEDAI \cite{razakarivony2016vehicle} and OGSOD \cite{wang2023category}, respectively. 
Considering that comparisons between models trained with different labels might lead to unfair results, this section also provided visualization cases to demonstrate the technical advantages of the proposed modules and model architecture.

\subsubsection{Comparison of Different Methods on DroneVehicle Dataset}
Table \ref{Comparison_of_Different_Methods_on_DroneVehicle_Dataset} presents the $AP_{50}$ results for each category and the $mAP_{50}$ across all categories on the DroneVehicle \cite{sun2022drone}.
The methods marked with `$^{\dagger}$' denote our reimplementation of other algorithms, whereas those without such annotations correspond to the original data presented in the paper.
Among the listed algorithms, our method achieves  $mAP_{50}$ of 78.36\%, 78.03\%, and 76.84\% under the supervision of VRSI labels, IRSI labels, and CMLF labels, respectively.
For a fair comparison, we compare the supervised results of the IRSI labels with the results of current multimodal oriented detectors.
The results of MO R-CNN significantly surpass those of newer single-spectrum oriented detection algorithms such as PKINet-S\cite{pu2023adaptive} and LSKNet-S\cite{li2023large}.
Furthermore, compared to multimodal detectors such as UA-CMDet\cite{sun2022drone}, Cascade-TSFADet\cite{yuan2022translation}, C2Former\cite{yuan2024c}, GAGTDet\cite{yuan2024improving}, DMM \cite{zhou2024dmm}, E2E-MFD\cite{zhang2024e2e} and DDCINet-RoI-Trans\cite{bao2025dual}, the $mAP_{50}$ is improved by 14.02\%, 4.97\%, 4.13\%, 3.83\%, 3.46\%, 0.43\%, 0.63 and 0.53\%, respectively.
We also explored the performance of Transformer-based or Mamba-based visual backbones in MO R-CNN.
As shown in Table \ref{Performance_of_Visual_Backbone_Based_on_Transformer_or_Mamba_in_MO_R-CNN}, compared to the Oriented R-CNN framework, Swin-S\cite{liu2021swin} and VMamba-S\cite{liu2024vmamba} demonstrated superior performance in MO R-CNN. 
For instance, our method brought at least a 1.17 $mAP_ {50}$ increase to VMamba-S\cite{liu2024vmamba}.

High metrics achieved under single-spectrum label supervision alone cannot adequately reflect the model's actual performance.
Fig. \ref{Visualization_of_Detection_Results_of_Recent_Methods_and_MO_R-CNN_on_DroneVehicle} provides a visualization of the ground truth labels and the detection results of the superior models.
From left to right, the panels show: VRSI labels, IRSI labels, results of the C2Former, results of the MO R-CNN trained with IRSI labels (Infrared-Model, I-Model), and results of the MO R-CNN trained with CMLF labels (CMLF-Model, C-Model).
Compared to C2Former\cite{yuan2024c}, our method demonstrates superior performance in both the I-Model and C-Model.
Due to differences in evaluation criteria, although the C-Model exhibits a slightly lower $mAP_{50}$ compared to the I-Model, it demonstrates greater advantages in cross-modal feature learning. In complex scenarios such as low-light conditions, the complementary label information from CMLF provides MO R-CNN with exceptional performance.

\begin{figure*}[!t]
	\centering
	\includegraphics[width=0.9\textwidth]{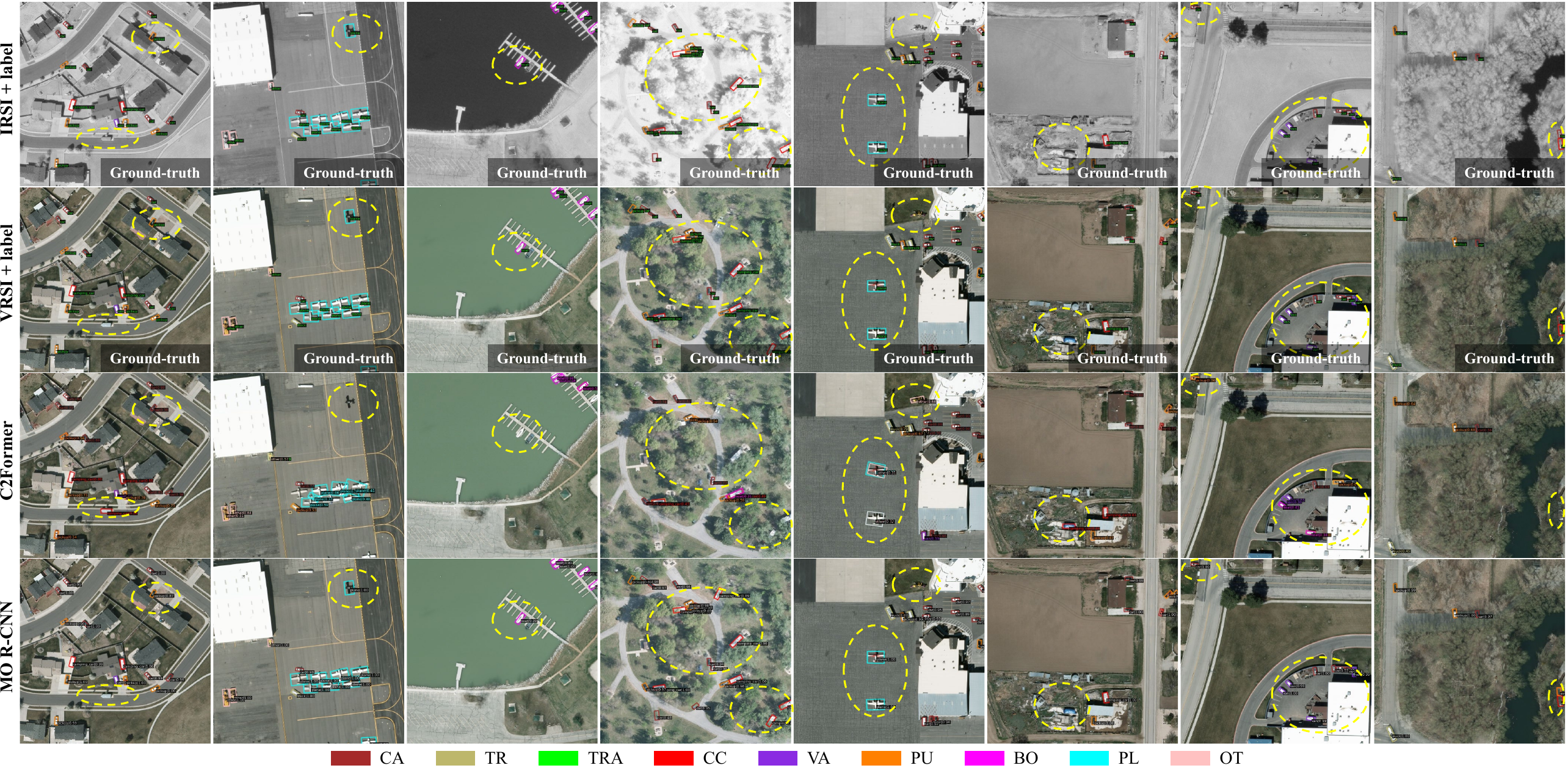}
	\caption{
		Visualization of detection results of recent methods and MO R-CNN on VEDAI. 
		From top to bottom, represent the results of IRSI label, VRSI label, results of C2Former, and results of MO R-CNN (C-M) respectively.
	}
	\label{Visualization_of_Detection_Results_of_Recent_Methods_and_MO_R-CNN_on_VEDAI}
\end{figure*}

\begin{figure*}[!t]
	\centering
	\includegraphics[width=0.9\textwidth]{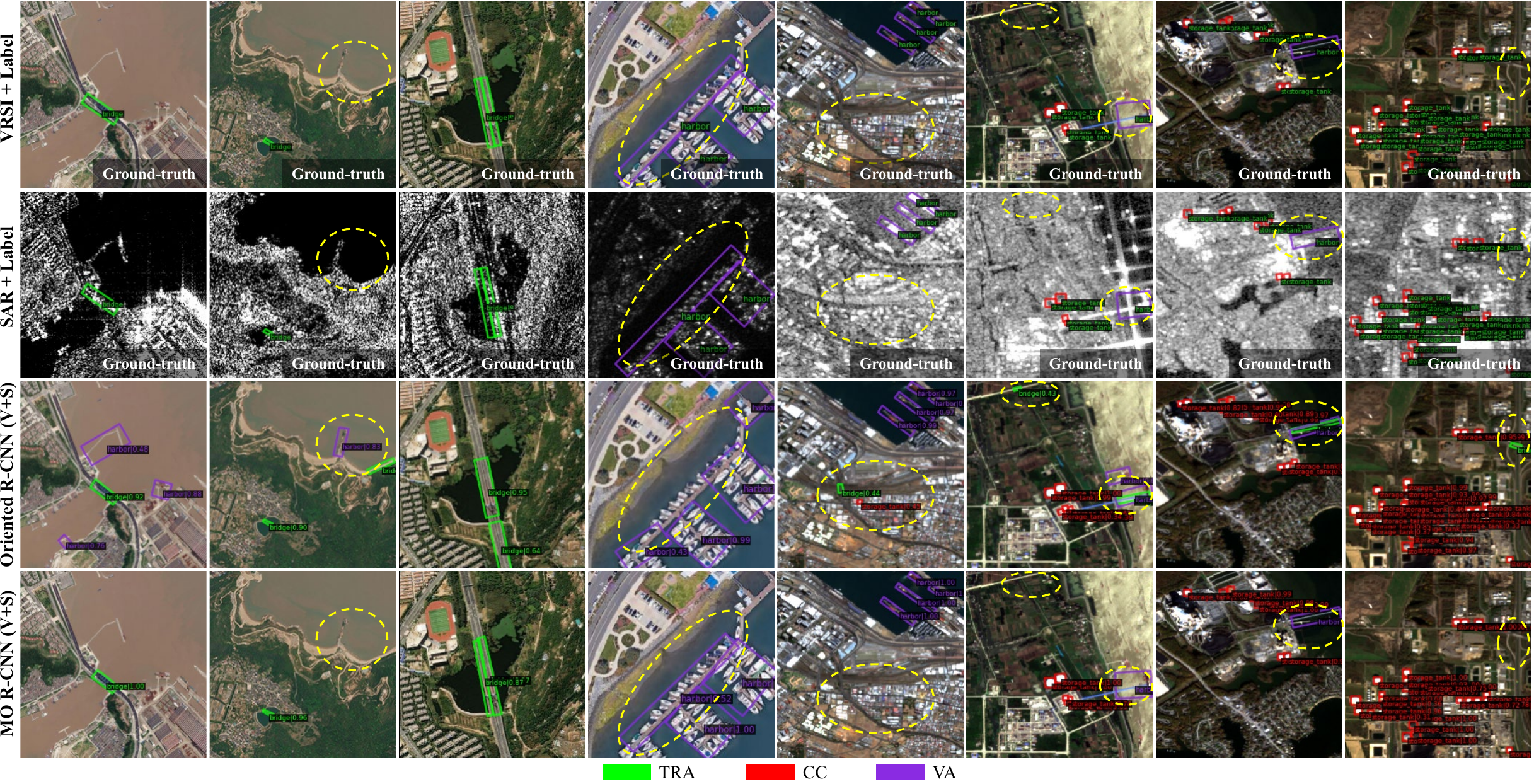}
	\caption{
		Visualization of detection results of Oriented R-CNN and MO R-CNN on OGSOD. 
		From top to bottom, represent the results of VRSI label, SAR label, results of Oriented R-CNN (VRSI+SAR), and results of MO R-CNN (VRSI+SAR) respectively.
	}
	\label{Visualization_of_Detection_Results_of_Oriented_RCNN_and_MO_R-CNN_on_OGSOD}
\end{figure*}

\begin{table}[!t]
	\renewcommand{\arraystretch}{1.0}
	\centering
	\footnotesize
	\caption{
		Comparison of MO R-CNN and Other Methods on OGSOD Dataset.
		Because the labels of the two modalities in the OGSOD dataset are exactly the same, CMLF labels were not used for training.
		The model was trained for 36 epochs.
	}
	\label{Comparison_of_Different_Methods_on_OGSOD_Dataset}
	\begin{tabular*}{\columnwidth}{@{\extracolsep{\fill}}@{\hspace{1pt}}l|c|c@{\hspace{1pt}}c@{\hspace{1pt}}c|l}
		\toprule
		\textbf{Method}	& \textbf{Modality} & \textbf{BR} & \textbf{ST} & \textbf{HA} &  \textbf{mAP}  \\
		
		\midrule
		DDCINet+O R-CNN\cite{bao2025dual}   				    & VRSI+SAR		& -    & -    & -    & 77.00 \\
		DDCINet+RoITrans\cite{bao2025dual}  				    & VRSI+SAR		& -    & -    & -    & 80.80 \\
		DDCINet+Swin-T\cite{bao2025dual}   	       				& VRSI+SAR		& -    & -    & -    & 81.10 \\
		\midrule
		Oriented R-CNN$^{\dagger}$\cite{xie2021oriented}  		& VRSI			& 75.2 & 65.5 & 86.5 & 75.76 \\
		Oriented R-CNN$^{\dagger}$\cite{xie2021oriented}  		& SAR			& 34.5 & 24.3 & 56.5 & 38.44 \\
		Oriented R-CNN$^{\dagger}$\cite{xie2021oriented}        & VRSI+SAR		& 77.5 & 69.1 & 89.7 & 78.78 \\
		\midrule
		MO R-CNN$^{\dagger}$ (ours)        		 			 	& VRSI+SAR		& 88.8 & 71.4 & 90.5 & \textbf{83.57} \\
		
		\bottomrule
	\end{tabular*}
\end{table}

\begin{table*}[!t]
	\renewcommand{\arraystretch}{1.0}
	\footnotesize
	\centering
	\caption{
		Ablation studies on the DroneVehicle dataset.
		These results indicate that each component contributes to the overall improvement in mAP.
	}
	\label{Ablation_studies_on_the_DroneVehicle_dataset}
	\begin{tabular*}{\textwidth}{@{\extracolsep{\fill}}l|lc|ccccc|cl}
		\toprule
		\textbf{Method} & \textbf{Module Combinatiion} & $\bm{\lambda}$, $\bm{\eta}$ & \textbf{CA} & \textbf{TR} & \textbf{BU} & \textbf{VA} & \textbf{FC} & \textbf{mAP@0.5}		 & \textbf{Params}  \\
		
		\midrule
		Baseline
		& -		                        & -         & 90.1 & 63.8 & 88.4 & 41.0 & 51.80 & 67.0  & 41.13 (Single-Modal) \\				
		
		\midrule
		\multirow{3}{*}{Baseline + HFEN}  
		& HDS-LSK						& -      	& 89.9  & 77.6 & 89.4 & 60.6 & 59.9 & 75.49 & 47.13 (Multi-Modal) \\
		& HDS-LSK + RFA					& -      	& 90.4  & 79.4 & 89.8 & 61.2 & 63.7 & 76.89 & 47.73 (Multi-Modal) \\
		& HDS-LSK + SMFF  				& -      	& 90.0  & 79.3 & 89.8 & 63.2 & 61.4 & 76.60 & 52.75 (Multi-Modal) \\
		& HDS-LSK + RFA + SMFF  		& -      	& 90.4  & 79.3 & 89.6 & 62.8 & 63.9 & 77.17 & 54.90 (Multi-Modal) \\
		
		\midrule
		\multirow{5}{*}{Baseline + HFEN + SMS}  
		& HDS-LSK + RFA + SMFF + SMS	& 1     	& 90.4  & 80.1 & 89.8 & 61.3 & 64.7 & 77.25 & 54.90 (Multi-Modal) \\
		& HDS-LSK + RFA + SMFF + SMS	& 0.5       & 90.4  & 80.1 & 89.7 & 61.9 & 64.1 & 77.24 & 54.90 (Multi-Modal) \\
		& HDS-LSK + RFA + SMFF + SMS	& 0.25      & 90.4  & 80.4 & 89.7 & 61.9 & 64.2 & 77.33 & 54.90 (Multi-Modal) \\
		& HDS-LSK + RFA + SMFF + SMS	& 0.125     & 90.4  & 80.3 & 89.8 & 62.2 & 64.7 & 77.47 & 54.90 (Multi-Modal) \\
		& HDS-LSK + RFA + SMFF + SMS	& 0.0625    & 90.4  & 79.7 & 89.8 & 65.8 & 64.5 & \textbf{78.03} & 54.90 (Multi-Modal) \\
		
		\bottomrule
	\end{tabular*}
\end{table*}

\subsubsection{Comparison of Different Methods on VEDAI Dataset}
Table \ref{Comparison_of_Different_Methods_on_VEDAI_Dataset} presents the $AP_{50}$ for each category and the $mAP_{50}$ across all categories on the VEDAI \cite{razakarivony2016vehicle}.
Since the two modalities in the VEDAI \cite{razakarivony2016vehicle} share identical labels, the CMLF label was not employed for training.
Among the listed algorithms, MO R-CNN achieves the highest $mAP_{50}$ of 94.2\% with $512 \times 512$ input images.
Under the same conditions (12 epoch), MO R-CNN surpasses C2Former\cite{yuan2024c}, CMAFF\cite{qingyun2022cross}, DMM\cite{zhou2024dmm}, Super YOLO(HBB)\cite{zhang2023superyolo}, DDCINet \cite{bao2025dual}, YOLOFIV \cite{wang2024yolofiv}, LF-MDet(HBB) \cite{sun2024low} and EMCFormer(HBB) \cite{wang2025emcformer} by $mAP_{50}$ margins of 38.0\%, 24.4\%, 18.6\%, 18.5\%, 10.9\%, 7.7\%, 7.1\% and 3.2\% respectively.

Fig. \ref{Visualization_of_Detection_Results_of_Recent_Methods_and_MO_R-CNN_on_VEDAI} provides visualizations of the ground-truth labels and model results.
C2Former\cite{yuan2024c} may exhibit missed detections or false positives in certain scenarios, while MO R-CNN demonstrates higher accuracy and robustness, particularly in complex environments such as residential areas and park scenes with tree occlusion.
Although VEDAI \cite{razakarivony2016vehicle} exhibits more diverse object categories and smaller-sized instances (with an size reduction of approximately 37\%) compared to DroneVehicle \cite{sun2022drone}, the MO R-CNN algorithm still demonstrates remarkable multi-scene generalization capabilities.

\subsubsection{Comparison of Different Methods on OGSOD Dataset}
Table \ref{Comparison_of_Different_Methods_on_DroneVehicle_Dataset} and Table \ref{Comparison_of_Different_Methods_on_VEDAI_Dataset} demonstrate the superior performance of our method in the VRSI+IRSI dual-modal configuration. 
Further experiments revealed that the approach is equally effective for the VRSI+SAR combination, as shown in Table \ref{Comparison_of_Different_Methods_on_OGSOD_Dataset}, where MO R-CNN achieved an mAP of 83.57 on OGSOD\cite{wang2023category}, surpassing the baseline and the latest method DDCINet by 4.79\% and 2.47\% respectively.
Fig. \ref{Visualization_of_Detection_Results_of_Oriented_RCNN_and_MO_R-CNN_on_OGSOD} demonstrates the improvements in object detection accuracy and robustness achieved by MO R-CNN with multimodal input compared to Oriented R-CNN with single-modal input.

\subsection{Ablation Studies}
This section conducts ablation experiments to validate the advantages of the proposed modules through both quantitative metrics and qualitative case studies.

\subsubsection{Quantitative Ablation Study}
We conducted our ablation experiments on the DroneVehicle \cite{sun2022drone}. 
To eliminate the influence of different detectors, we adopted the classic Oriented R-CNN as the baseline and modified it into a dual-stream oriented detector.
As shown in Table \ref{Ablation_studies_on_the_DroneVehicle_dataset}, incorporating the HDS-LSK resulted in a 8.49\% increase in mAP. 
When the RFA and SMFF modules are applied individually, they improve the baseline mAP by 1.4\% and 1.1\%, respectively, demonstrating the effectiveness of RFA and SMFF in processing multi-modal remote sensing images.
When the RFA and SMFF modules are used together, the mAP improves by 1.68\% compared to the baseline, demonstrating their complementary effects and mutual enhancement in handling multi-modal remote sensing images.
When performing the ablation study on SMS, we gradually set $\bm{\lambda}$ and $\bm{\eta}$ to 1, 0.5, 0.25, 0.125, 0.0625 to further investigate the impact of the supervision loss, and obtained the optimal loss balancing weight of 0.0625.
Integrating all the modules, the model achieved its best performance with an mAP of 78.03\%.
These results indicate that each component contributes to the overall improvement in mAP.

\begin{figure}[!t]
	\centering
	\includegraphics[width=1.0\columnwidth]{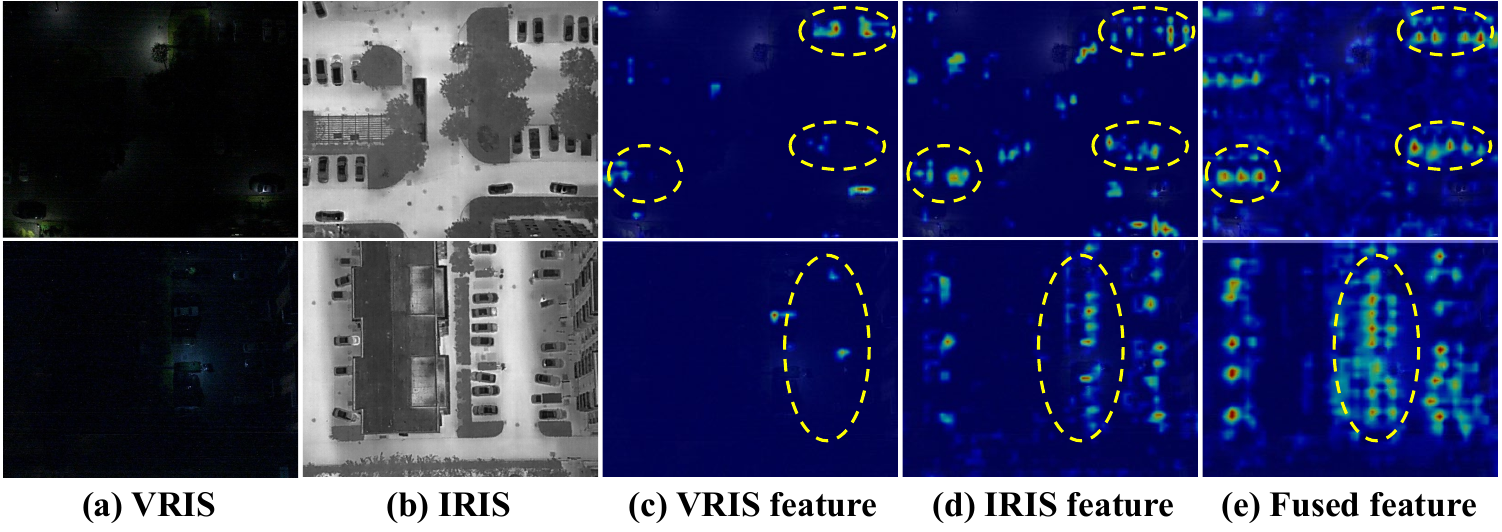}
	\caption{
		Visualization of modal fusion. 
		The fused feature on the far right is more comprehensive and robust in target representation compared to the unimodal feature representations of VRIS and IRIS.
	}
	\label{Visualization_of_Modal_Fusion}
\end{figure}

\subsubsection{Qualitative Ablation Study}
The SMFF demonstrated exceptional fusion performance during the feature fusion process.
Fig. \ref{Visualization_of_Modal_Fusion} demonstrated the unimodal features (VRIS and IRIS) and fused features of three image pairs. The fused features on the far right exhibited more comprehensive and robust target representations compared to their unimodal counterparts.
Compared to static fusion methods, the SMFF improved both adaptability and performance of multimodal remote sensing detection models in complex scenarios (e.g., day-night environments) by preserving modality specificity while achieving optimal coupling of heterogeneous features through learnable offset fields.

\begin{figure}[!t]
	\centering
	\includegraphics[width=1.0 \columnwidth]{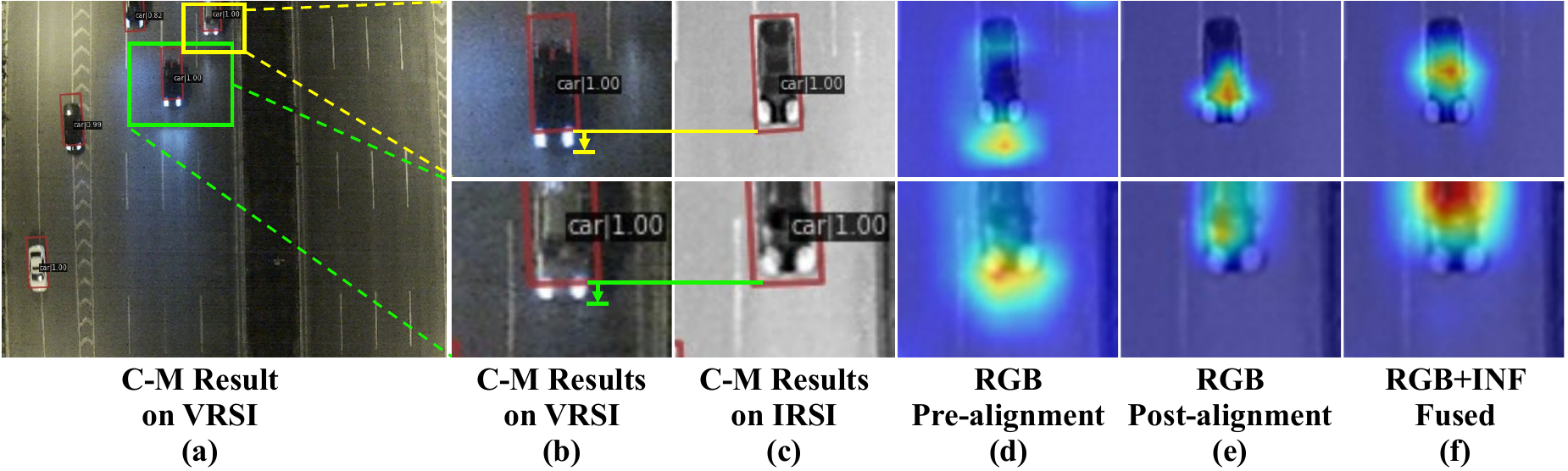}
	\caption{
		Visualization of modal alignment with small spatial deviations. 
		When the target position offset is small, the alignment mechanism of SMFF can effectively and precisely align the visible light features with the infrared object.
	}
	\label{Visualization_of_Modal_Alignment1}
\end{figure}

\begin{figure}[!t]
	\centering
	\includegraphics[width=1.0\columnwidth]{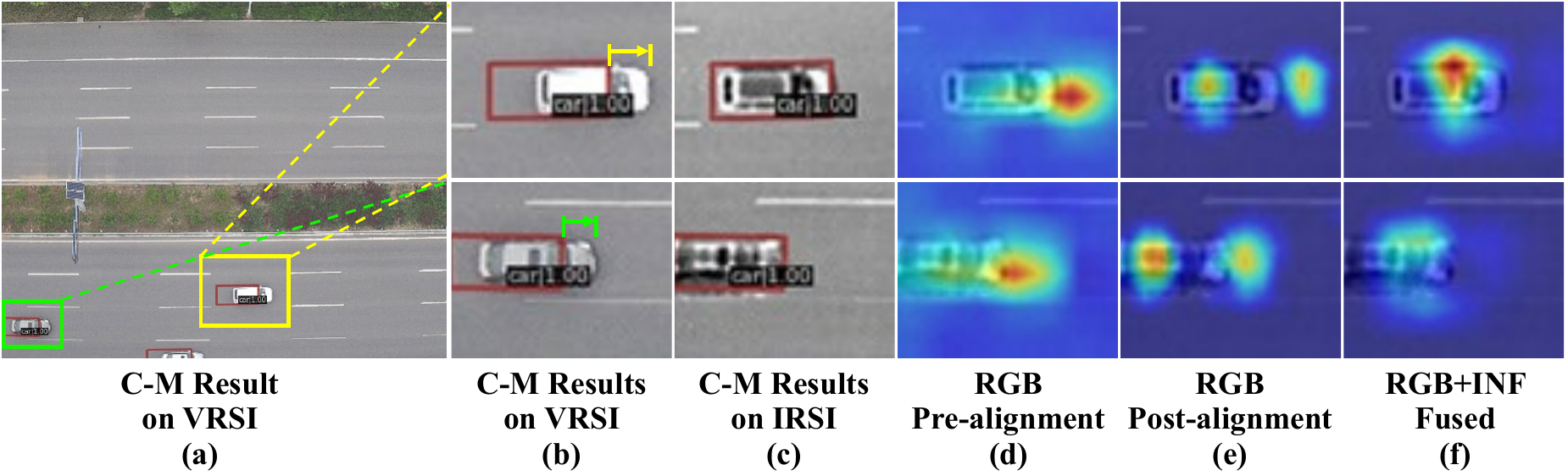}
	\caption{
		Visualization of modal alignment with large spatial deviations. 
		In cases where the target position offset is large, although the alignment module may result in the truncation of visible light features, the final fused feature can still maintain good representational capability through the subsequent attention fusion mechanism.
	}
	\label{Visualization_of_Modal_Alignment2}
\end{figure}

The feature fusion performance of the SMFF was significantly dependent on its efficient feature alignment mechanism.
Fig. \ref{Visualization_of_Modal_Alignment1} and Fig. \ref{Visualization_of_Modal_Alignment2} presented four instances of multimodal feature alignment. (a) and (b) represented the visualizations of the CMLF-trained model results (CMLF Model, C-M) on VRSI, while (c) demonstrated its corresponding results on IRSI. (d) and (e) corresponded to the pre-alignment and post-alignment visible feature maps in SMFF with IRSI as the background, whereas (f) displayed the fused feature map output by SMFF under the same IRSI background.
Fig. \ref{Visualization_of_Modal_Alignment1}(d) and (e) demonstrated that the alignment mechanism of SMFF was capable of effectively achieving precise alignment between visible features and infrared objects when positional offsets were minimal.
Fig. \ref{Visualization_of_Modal_Alignment2}(d) and (e) demonstrated that under conditions of large positional offsets, although the alignment module might induce truncation of visible features, the final fused features maintained robust representation capabilities through the subsequent attention-based fusion mechanism.
These examples demonstrated the significant superiority of the SMFF module in feature alignment.

\begin{table}[!t]
	\renewcommand{\arraystretch}{1.0}
	\footnotesize
	\centering
	\caption{
		comparisonn of our MO R-CNN with the latest sota method C2Former in terms of the number of model parameters.
	}
	\label{comparison_of_our_MO_R-CNN_with_the_latest_sota_method_C2Former_in_terms_of_the_number_of_model_parameters}
	\begin{tabular*}{0.95\columnwidth}{@{\extracolsep{\fill}}c|c|c|c}
		\toprule
		\textbf{Method} & \textbf{Params} & \textbf{Max. input} & \textbf{mAP@0.5} \\
		
		\midrule
		C2Former	 			& 100.80M				& $928 \times 928$ 		& 74.20   		 \\
		MO R-CNN				& 54.90M(45.9M $\downarrow$)		& $ 1728 \times 1728$ 	& 78.03(3.83$\uparrow$)   \\
		
		\bottomrule
	\end{tabular*}
\end{table}

\begin{figure}[!t]
	\centering
	\includegraphics[width=0.95\columnwidth]{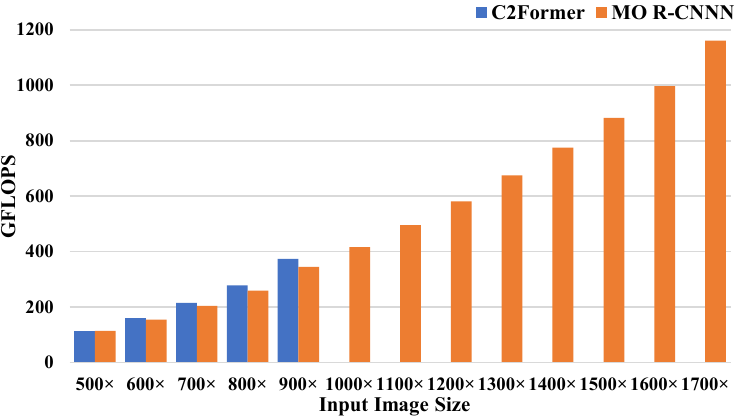}
	\caption{
		Comparison of GFLOPS for the model with different input image sizes. 
		In contrast, although the GFLOPS of C2Former initially approached our model on smaller image sizes, the memory usage of C2Former significantly increased as the input image size increased. 
		This results in a maximum inference size supported by C2Former of only $928 \times 928$ on a 24 GB NVIDIA 3090 GPU, while our model can handle inputs of $1728 \times 1728$, which is approximately 3.47 times larger.
	}
	\label{Comparison_of_GFLOPS_for_the_model_with_different_input_image_sizes}
\end{figure}

\subsection{Computational Cost Comparison}
As shown in Table \ref{Ablation_studies_on_the_DroneVehicle_dataset}, compared to the single-branch unimodal baseline model, our proposed dual-branch multimodal network architecture would theoretically double the parameter. 
However, due to the carefully designed lightweight HDS-LSK module, the actual parameter increase is only 13.77 M, enabling the dual-branch structure to maintain a parameter scale comparable to that of the single-branch baseline. 
This efficient parameter design makes our method significantly outperform recent mainstream approaches in terms of model parameters, as shown in Table \ref{Comparison_of_Different_Methods_on_DroneVehicle_Dataset}.

Table \ref{comparison_of_our_MO_R-CNN_with_the_latest_sota_method_C2Former_in_terms_of_the_number_of_model_parameters} presents a comparative analysis between our MO R-CNN and C2Former with respect to model parameter count.
MO R-CNN reduces parameter to just 54.90M (a decrease of approximately 45.54\%) while also achieving the highest detection performance.
Additionally, as illustrated in Fig. \ref{Comparison_of_GFLOPS_for_the_model_with_different_input_image_sizes}, the GFLOPS of our model exhibit an approximately linear increase with growing input image sizes.
In comparison, while C2Former's GFLOPS remain comparable to our model at smaller image sizes, its memory consumption escalates significantly with increasing input dimensions. Consequently, on a 24 GB NVIDIA 3090 GPU, C2Former supports a maximum inference size of only $928 \times 928$, whereas our model can handle $1728 \times 1728$  inputs — approximately 3.67 times larger.
This experimental result demonstrates that our model exhibits superior capability in processing remote sensing images.

\section{Conclusion}
In this paper, we proposed a lightweight multispectral convolutional neural network (MO R-CNN), a framework for multispectral oriented object detection in remote sensing, which includes HFEN, SMS, and CMLF.
HFEN achieved differentiated feature extraction across multiple inputs through HDS-LSK, aligned and fused cross-modal features via SMFF, and enhanced multi-scale features using RFA, significantly reducing spectral-specific information attenuation.
SMS is employed solely during the training phase without affecting the inference stage, constraining multi-scale features while endowing the model with the capability to learn from multiple modalities.
CMLF fuses bimodal labels according to specific rules, marking the first non-model improvement approach to alleviate the bimodal label inconsistency issue in real-world scenarios.
Extensive experiments conducted on three challenging datasets demonstrated the generalization capability of MO R-CNN, establishing a new baseline for multi-spectral oriented detection.
In the future, we will explore the application of our method in broader scenarios.


%




\ifCLASSOPTIONcaptionsoff
  \newpage
\fi



\bibliographystyle{IEEEtran}	
\bibliography{mo_r-cnn.bib}   

\end{document}